\documentclass[sigconf]{acmart}

\AtBeginDocument{%
  \providecommand\BibTeX{{%
    \normalfont B\kern-0.5em{\scshape i\kern-0.25em b}\kern-0.8em\TeX}}}

\copyrightyear{2022} 
\acmYear{2022} 
\setcopyright{acmcopyright}
\acmConference[CCS '22]{Proceedings of the 2022 ACM
SIGSAC Conference on Computer and Communications Security}{November 7--11,
2022}{Los Angeles, CA, USA}
\acmBooktitle{Proceedings of the 2022 ACM SIGSAC Conference on Computer and
Communications Security (CCS '22), November 7--11, 2022, Los Angeles, CA, USA}
\acmPrice{15.00}
\acmDOI{10.1145/3548606.3559392}
\acmISBN{978-1-4503-9450-5/22/11}

\usepackage{times}  % DO NOT CHANGE THIS
\usepackage{courier}  % DO NOT CHANGE THIS
\usepackage{url}  % DO NOT CHANGE THIS
\usepackage{graphicx} % DO NOT CHANGE THIS
\usepackage{xspace}
\usepackage{amsmath}
\usepackage{amsfonts}
\usepackage{amsthm}
\usepackage{booktabs}
\usepackage{graphicx}
\usepackage{caption}
\usepackage{subcaption}
\usepackage[abbreviations]{foreign}
\usepackage{bbding}

\usepackage{caption} % DO NOT CHANGE THIS AND DO NOT ADD ANY OPTIONS TO IT
\usepackage[linesnumbered,algoruled,boxed,lined]{algorithm2e}
\usepackage[abbreviations]{foreign}
\usepackage{multirow}
\usepackage{newfloat}
\usepackage{listings}

\usepackage{hyperref}

\DeclareMathOperator*{\argmin}{argmin}
\DeclareMathOperator*{\argmax}{argmax}
\DeclareMathOperator*{\ST}{\mathcal{S}}
\DeclareMathOperator*{\Var}{\mathrm{Var}}
\DeclareMathOperator*{\Cor}{\mathrm{Cor}}

\DeclareMathOperator*{\N}{\mathbb{N}}

\theoremstyle{Proposition}
\newtheorem{proposition}{Proposition}
\newtheorem{theorem}{Theorem}

\newtheorem{definition}{Definition}

\usepackage{xcolor}

\newcommand{\paragraphbe}[1]{\smallskip\noindent{\bf {#1}.}~}

\SetKwInOut{Parameter}{Parameters}

\begin{document}
\title{``Is your explanation stable?'': A Robustness Evaluation Framework for Feature Attribution} 
\author{Yuyou Gan}
\authornote{Yuyou Gan and Yuhao Mao contributed equally. Xuhong Zhang and Shouling Ji are the corresponding authors. }
\affiliation{%
 \institution{Zhejiang University}
 \country{}
}
\email{ganyuyou@zju.edu.cn}

\author{Yuhao Mao}
\authornotemark[1]
%\author{Shouling Ji}
\affiliation{%
 \institution{Zhejiang University, ETH Zürich}
 \country{}
% \institution{Alibaba-Zhejiang University Joint Institute of Frontier Technologies}
}
\email{yuhaomao@zju.edu.cn}

\author{Xuhong Zhang}
\authornotemark[0 \Envelope]
\affiliation{%
 \institution{Zhejiang University}
 \country{}
}
\email{zhangxuhong@zju.edu.cn}

\author{Shouling Ji }
\authornotemark[0 \Envelope]
\affiliation{%
 \institution{Zhejiang University}
 \country{}
}
\email{sji@zju.edu.cn}

\author{Yuwen Pu}
\affiliation{%
 \institution{Zhejiang University}
 \country{}
}
\email{yw.pu@zju.edu.cn}

\author{Meng Han}
\affiliation{%
 \institution{Zhejiang University}
 \country{}
}
\email{mhan@zju.edu.cn}

\author{Jianwei Yin}
\affiliation{%
 \institution{Zhejiang University}
 \country{}
}
\email{zjuyjw@zju.edu.cn}

\author{Ting Wang}
\affiliation{\institution{The Pennsylvania State University}
\country{}
}
\email{inbox.ting@gmail.com}

\begin{abstract}

Neural networks have become increasingly popular. Nevertheless, understanding their decision process turns out to be complicated. One vital method to explain a models' behavior is feature attribution, \ie, attributing its decision to pivotal features. Although many algorithms are proposed, most of them aim to improve the faithfulness (fidelity) to the model. However, the real environment contains many random noises, which may cause the feature attribution maps to be greatly perturbed for similar images. More seriously, recent works show that explanation algorithms are vulnerable to adversarial attacks, generating the same explanation for a maliciously perturbed input. All of these make the explanation hard to trust in real scenarios, especially in security-critical applications.

To bridge this gap, we propose \emph{Median Test for Feature Attribution} (MeTFA) to quantify the uncertainty and increase the stability of explanation algorithms with theoretical guarantees. MeTFA is method-agnostic, \ie, it can be applied to any feature attribution method. MeTFA has the following two functions: (1) examine whether one feature is significantly important or unimportant and generate a MeTFA-significant map to visualize the results; (2) compute the confidence interval of a feature attribution score and generate a MeTFA-smoothed map to increase the stability of the explanation.
Extensive experiments show that MeTFA improves the visual quality of explanations and significantly reduces the instability while maintaining the faithfulness of the original method. To quantitatively evaluate MeTFA's faithfulness and stability, we further propose several robust faithfulness metrics, which can evaluate the faithfulness of an explanation under different noise settings. Experiment results show that the MeTFA-smoothed explanation can significantly increase the robust faithfulness. In addition, we use two typical applications to show MeTFA's potential in the applications. First, when being applied to the SOTA explanation method to locate context bias for semantic segmentation models, MeTFA-significant explanations use far smaller regions to maintain 99\%+ faithfulness. Second, when testing with different explanation-oriented attacks, MeTFA can help defend vanilla, as well as adaptive, adversarial attacks against explanations.

\end{abstract}

\begin{CCSXML}
<ccs2012>
   <concept>
       <concept_id>10010147.10010257.10010293.10010294</concept_id>
       <concept_desc>Computing methodologies~Neural networks</concept_desc>
       <concept_significance>500</concept_significance>
       </concept>
 </ccs2012>
\end{CCSXML}

\ccsdesc[500]{Computing methodologies~Neural networks}

\keywords{Explanation of AI; Robustness of Explanation; Hypothesis Testing}

\maketitle

\vspace{-1mm}
\section{Introduction}

The extraordinary performance of deep neural networks (DNNs) has led us to the era of deep learning \cite{arp2014drebin}\cite{he2016deep}\cite{sutskever2014sequence}\cite{9409758}. While these networks achieve human-level performance on various tasks, there are many security crises in DNNs \cite{li2022seeing}\cite{pang2022security}\cite{fu2022label}\cite{mao2022transfer}, which  
% they remain incomprehensible, or so-called ``black-box'', to humans. This undesired property
prevent users from fully trusting the models' output, raising concerns for sensitive applications like autonomous driving. Many works are proposed to make DNNs more safe and reliable \cite{zheng2022neuronfair}\cite{du2021cert}\cite{li2020textshield}, one of which is to explain the action of model. By interpreting the black-box model, we can detect biases and anomalies, and gain insights to improve the model. %\topic{Explanation is important.}

Feature attribution, which explains the model at instance level, is a popular method to explain neural networks \cite{Petsiuk_2021_CVPR}\cite{wagner2019interpretable}\cite{Gu_2021_CVPR}\cite{Lee_2021_CVPR}. These explanations give each feature a feature score representing its contribution to the model's output. Features with high contribution scores support the decision of the model, and thus we call them \emph{supportive features}. For instance, in the image domain, attribution map is a common form of feature attribution. From the attribution map, users can see which parts of the image are relied on by the model to make the prediction. In addition, feature attribution methods help to create better models. For example,
recent works use feature attribution to debug errors \cite{guo2018lemna}, detect adversarial examples \cite{zhang2020interpretable}, and check context bias \cite{hoyer2019grid}. %\topic{Feature attribution is a popular explanation.}

\begin{figure}[t]
     \centering
     \includegraphics[width=0.7\columnwidth]{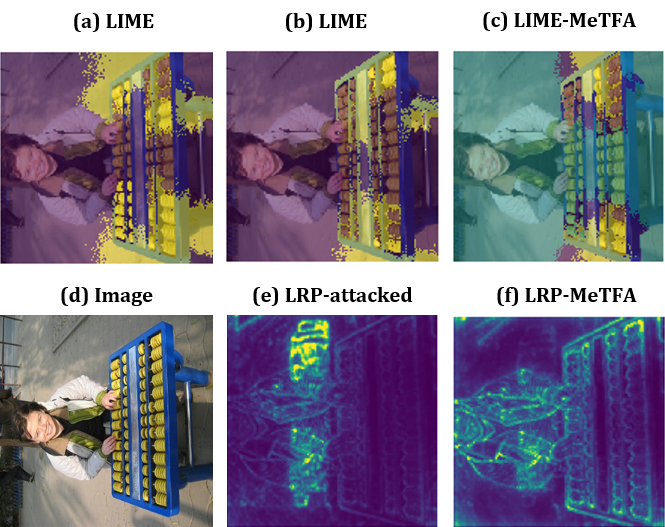}
     \caption{Example explanations on VGG16. (a) and (b) are the explanations by two independent runs of LIME \cite{ribeiro2016should}, a black-box explanation method. (c) is the MeTFA-significant LIME explanation, where the yellow area is significantly important, the green area is significantly unimportant, and the rest region is insignificant. (d) is the seed  image of manipulation attack \cite{dombrowski2019explanations}, an attack that keeps the model's prediction unchanged but manipulates the explanation of the prediction. (e)Attack\cite{dombrowski2019explanations} for LRP\cite{bach2015pixel} shows the explanation of the LBP for the adversarial image generated by manipulation attack from (d), where the explanation is totally irrelevant to the prediction abacus. (f) is the MeTFA-smoothed explanation of the same adversarial image, where a plausible explanation is recovered. }
     
     \label{Fig_intro}
\end{figure}
                      
However, the feature attribution methods used to explain neural networks are still facing a reliability crisis. The feature attribution methods can be roughly classified into two categories: black-box methods and white-box methods. Black-box methods only have access to the input and output of a network. Generally, such methods apply various forms of sampling, which leads to uncertainty in the explanation \cite{guo2018lemna}\cite{petsiuk2018rise}\cite{ribeiro2016should}. For example, as shown in Figure \ref{Fig_intro} (a) and (b), LIME \cite{ribeiro2016should} gives different attribution maps for the same image in independent runs. On the contrary, white-box methods mainly use deterministic information, such as gradient \cite{bach2015pixel}\cite{Gradient} and activation maps \cite{zhou2016learning}\cite{wang2020score}, to generate attribution maps, thus guarantee a deterministic explanation for the same input. However, they still have problems due to instability. For example, gradient-based methods produce drastically different attribution maps for similar inputs \cite{smilkov2017smoothgrad}; optimization-based methods sometimes generate nonsensical or unexpected attribution maps due to the instability of the non-robust features \cite{fong2017interpretable}. 
% These uncertainties greatly affect the reliability of the explanation, making it unsuitable for areas that require stable explanations, such as autonomous driving.  
Moreover, explanations with high sensitivity may be more vulnerable to adversarial attacks \cite{generalSG}. For example, in Figure \ref{Fig_intro} (d) and (e), by adding human-imperceptible noise, the attacker can manipulate an attribution map arbitrarily. These phenomena greatly reduce our trust in explanation algorithms. %\topic{Attribution maps contain unaddressed uncertainty.}

As illustrated above, existing feature attribution methods give uncertain results due to the sampling process or the non-robust features. Therefore, to make the explanations reliable, a method to reduce and quantify the uncertainty involved in the explanation is required.
% To tackle these problems, a promising way is to sample around the original data and take all of the explanations of these samples into consideration instead of only the explanation of the original data.
To increase the stability of the explanations, a promising way is to sample inputs from the neighborhood of the original input and average all these explanations. When the explanation is Gradient \cite{Gradient}, this method is known as SmoothGrad \cite{smilkov2017smoothgrad}.
% There are some attempts to increase the stability of the explanations.To increase the stability of gradient-based explanations, SmoothGrad \cite{smilkov2017smoothgrad} samples inputs from the neighborhood of the original input and averages over these explanations to produce a more robust explanation.
Therefore, by the law of large numbers, the smoothed explanation essentially converges to the expectation of the distribution of the neighborhood explanations of the original input, thus mitigating the uncertainty to some degree.

Although this method, along with its modification \cite{yeh2019fidelity}, are designed only to produce stable results, it can be further extended to quantify the uncertainty of the explanation by computing the standard deviation of the samples.  Specifically, by the central limit theorem, they can be extended to use the mean and standard deviation to derive an asymptotic confidence bound of the explanation. Details on how to derive the bound are included in Section \ref{sec:discuss}.
% \footnote{To distinguish the extended version with the original SmoothGrad, we call the extension by ``SmoothExp'' in the paper. The details about how to derive the confidence bounds are included in Section \ref{sec:related work}.} 
However, the correctness of these bounds requires the law of large numbers, \ie, they are only valid when the number of samples is extremely large. This brings heavy computational overheads because sampling explanations is computationally expensive. In addition, when the number of samples is small,
 the mean value is sensitive to abnormal extreme values. Since the underlying distribution of explanations is unknown, it is difficult to get any theoretical guarantees for the confidence bounds with such methods when the number of samples is small. Therefore, how to efficiently quantify the uncertainty of the explanations remains unresolved.

\vspace{-0.5mm}

\paragraphbe{Our designs}
To overcome the above challenges, we propose \emph{Median Test for Feature Attribution} (MeTFA). Instead of generating a single score for each feature, MeTFA takes a novel perspective: a reliable explanation should include the attribution score map together with the confidence interval and the significance of the scores. The core idea of MeTFA is to sample around the original data and conduct the hypothesis test on the samples' explanations to establish theoretical guarantees. To tackle the problem of unknown distribution and the impact of the abnormal extreme values, MeTFA approximates the median of the explanation distribution instead of the expectation. In this way, MeTFA converts the unknown distribution to a Bernoulli distribution on a specific statistics without approximating the normal distribution by the central limit theorem, thus allowing it to get exact confidence bounds with a small number of samples.
% Instead of approximating the expectation of the distribution, MeTFA approximates the median of the distribution. This is because hypothetical test needs a known distribution and MeTFA can convert the unknown distribution to the Bernoulli distribution by assuming that the median is above or below a certain value. In this way, we can establish theoretical guarantees for approximation by constructing hypothesis tests. 
MeTFA considers the following two scenarios: (1) the users only want to know which features are significantly important or unimportant with regard to a threshold of the feature score, and (2) the users wish to know the confidence bounds for the feature score. The first scenario is more general because many explanation algorithms, such as LIME \cite{ribeiro2016should}, only provide a discrete binary score as the feature attribution, while the second scenario requires a continuous feature score in the range of $[0,1]$.
For the first scenario, given a user-interested feature score threshold and a confidence level $\alpha$, we design the one-sided MeTFA to generate the MeTFA-significant map. This map shows whether one feature is significantly important (the median of the explanation distribution is higher than the threshold) or unimportant (the median of the explanation distribution is lower than the threshold) with regard to the confidence level $\alpha$. For the second scenario,  we design the two-sided MeTFA to compute the median's $\alpha$-confidence interval for the features. Further, by averaging the sampled feature scores in the confidence interval, MeTFA generates the MeTFA-smoothed explanation as the approximation of the median. This is different from SmoothGrad which averages over all the sampled scores. 
In Figure \ref{Fig_intro} (c) and (f), the results of LIME \cite{ribeiro2016should} with one-sided MeTFA and LRP \cite{bach2015pixel} with two-sided MeTFA show that MeTFA can reveal the significantly important features and defend against the explanation-oriented attacks. In addition, we prove by theoretical and empirical findings that the variance of the MeTFA-smoothed explanation shrinks to $0$, which suggests the correctness of MeTFA. Moreover, the variance shrinks with a same or faster speed than SmoothGrad, which suggests that MeTFA-smoothed explanations are more stable and efficient.

\vspace{-0.5mm}
\paragraphbe{Evaluations}
In the image domain, we evaluate our method on representatives of the four main types of explanation methods: (1) gradient-based method: Gradient \cite{Gradient}, (2) sample-based method: RISE \cite{petsiuk2018rise} and  LIME \cite{ribeiro2016should}, (3) optimization-based method: IGOS \cite{qi2019visualizing}, and (4) activation-based method: ScoreCAM \cite{wang2020score}. We use the popular metrics \emph{insertion}, \emph{deletion} \cite{petsiuk2018rise} and \emph{overall} \cite{zhang2021group} as the faithfulness  metrics and \emph{standard deviation} (std) of feature scores as the stability  metric. Further, we propose the \emph{robust insertion}, \emph{robust deletion} and \emph{robust overall} metrics to measure the ability of explanation to locate robust features.  The results show that MeTFA slightly affects the faithfulness for the orignal input but significantly increases the robust faithfulness. In addition, we show that MeTFA is better than SmoothGrad in stability.
% the feature scores in the MeTFA-smoothed explanation converge with a same or faster speed and experiments show that MeTFA-smoothed explanations are more stable and converge faster. 
In the NLP domain, we evaluate MeTFA with LEMNA \cite{guo2018lemna}, the SOTA explanation alogorithm targeting RNN. We use \emph{feature deduction test}, \emph{feature augmentation test} and \emph{synthetic test} proposed in \cite{guo2018lemna} as the faithfulness metrics and take \emph{std} and the \emph{overlap of top n features} as the stability evaluation metrics. Similar to the image domain, we propose three corresponding robust faithfulness metrics. Experiment results show that MeTFA can improve all the faithfulness, stability and robust faithfulness metrics for LEMNA.

\vspace{-0.5mm}
\paragraphbe{Applications}
To illustrate the potential of MeTFA in practice, we apply one-sided MeTFA and two-sided MeTFA, respectively, to two applications closely related to security: detecting the context bias in the semantic segmentation and defending the explanation-oriented adversarial attack. We apply the one-sided MeTFA to GridSaliency \cite{hoyer2019grid}, the SOTA method to locate context bias for semantic segmentation models. The results show that, in the environment with common noise distributions, MeTFA can greatly reduce the context bias region while maintaining 99\%+ faithfulness, which suggests the potential of MeTFA to detect context bias in the real world. Moreover, we demonstrate that two-sided MeTFA can defend both the attacks that produce wanted explanations while keeping model's predictions unchanged and the attacks that produce unchanged explanations for wanted predictions. 
% In addition, we observe that the (explanation, MeTFA-smoothed explanation) pair behave differently for clean images and adversarial examples, which may assist in detecting adversarial examples.

\vspace{-0.5mm}
\paragraphbe{Contributions}
(1) We propose a novel perspective: a reliable explanation should include not only the attribution score map but also the confidence interval and the significance of the scores. (2) We propose a framework MeTFA to quantify the uncertainty and increase the stability of the feature attribution algorithm with theoretical guarantees. (3) We propose a series of robust metrics which consider the neighborhood of the input instead of a single point. Experimental results show that MeTFA increases the stability while maintaining the faithfulness. (4) The application of MeTFA on detecting the context bias in semantic segmentation and defending against the adversarial examples with the explanation-oriented attack shows its great potential in real practice.

\vspace{-2mm}
\section{Related Work}\label{sec:related work}

\textit{Visual explanation}. Visual explanation methods can be divided into white-box methods and black-box methods. White-box explanations are free to use all the information about the model, such as architecture and parameter. They can be roughly categorized into three groups: gradient-based, activation-based and optimization-based. Gradient-based methods \cite{baehrens2010explain} \cite{Gradient} use gradient information to generate the feature importance of pixels, known as saliency maps. These methods are typically fast but may render volatile explanations due to the sensitivity of the gradients of input samples \cite{smilkov2017smoothgrad}. Activation-based methods \cite{zhou2016learning} \cite{selvaraju2017grad} \cite{wang2020score} address this problem by using the activations of convolutional layers instead. They apply upscaled linear combination of some layers' output as the explanation and find that it usually highlights the region of the correct object and is more stable. Optimization-based methods \cite{zhang2021novel} \cite{wagner2019interpretable} \cite{fong2017interpretable} \cite{qi2019visualizing} do not generate feature importance for all the pixels but highlight a small area of interest using optimization. They typically generate high-quality explanations but are much slower, as the optimization process requires multiple forward and backward propagations. Black-box explanations \cite{petsiuk2018rise} \cite{ribeiro2016should} \cite{petsiuk2021black} only require the access to the input and output of the model. They randomly sample some features, modify these features (\eg LIME sets them to $0$), put the modified inputs into the model and explain using the outputs (\eg LIME uses a linear model to fit the outputs and the modified inputs). It can be found that MeTFA and sample-based explanation methods use sampling in different ways and for different purposes. MeTFA uses the common noise in real life to sample inputs around the original input and then conduct hypothetical testing on their explanations to increase the stability of explanation methods in the real world.
% Typical processing includes using the predicted score of the model as the weight of summation \cite{wang2020score} \cite{petsiuk2018rise} and fitting a surrogate model \cite{ribeiro2016should} \cite{guo2018lemna} \cite{ribeiro2018anchors}.

% \subsection{Visual explanation}
% We can separate visual explanation method into two broad classes based on whether it is necessary to obtain model internal information (such as gradients, features, etc.). 

% White-box methods using gradient of target class(\TBD{gradient-based method}), activation maps(\TBD{activation-based method}) or defining an optimization function(\TBD{optimization method}) to generate a saliency map. Each method has its own characteristics. Gradient-based method are fast and intuitive however tend to generate lower quality saliency maps. Activation-based method generate saliency maps by linearly combining the activation maps after some layer and then upsampling to the the same resolution of the original input image. Optimization-based methods are to find the minimum region to find  a minimal subset of sufficient evidence. Optimization-based methods are more flexible to activation-based methods and more high-quality than gradient-based methods but slower. 

% Black-box methods query the model to generate a surrogate model locally(\TBD{Anchor and LIME} or generate a saliency map by combining candidate maps(\TBD{medical and rise}). All of them need to sample some data and utilize the information returned by the model to achieve explanation. Black-box method are always model-free but slow and the results are uncertain.

\textit{Evaluation of visual explanation.} {\v{S}}ikonja \etal \cite{Robnik-sikonja2018} summarized the required properties of explanations. Some of them are: (1) faithfulness to the model, (2) stability of the explanation when the input is slightly perturbed, and (3) comprehensibility of the explanation to humans. Unfortunately, current explanation algorithms are not satisfactory at these properties. Ghorbani \etal \cite{DBLP:conf/aaai/GhorbaniAZ19} found that explanations could be largely affected by adversarial perturbations which do not change the model's prediction. In addition, Kindermans \etal \cite{DBLP:series/lncs/KindermansHAASDEK19} found that for two networks with provably same explanations, the explanations for the two networks produced by current explanation algorithms are different. Moreover, Adebayo \etal \cite{DBLP:conf/nips/AdebayoGMGHK18} found that when the network is gradually randomized, many algorithms do not produce randomized explanations. Instead, they highlight ``edge pixels'', which is similar to edge detectors. By evaluating recent explanation algorithms, these works show that there is a large gap to fulfill in the explanation algorithms.

\textit{Post hoc improvement on the explanation.} There are some attempts aiming at improving the stability of explanations by post hoc improvements. They are built on a stability assumption that the explanations should not vary greatly for similar inputs. Smilkov \etal \cite{smilkov2017smoothgrad} first showed that the gradient-based explanations were vulnerable to small perturbation to the input. They found by experiment that adding Gaussian noises to the input and averaging their explanations would provide an explanation with better visual quality. This method is called \emph{SmoothGrad}. To understand why SmoothGrad works, Yeh \etal \cite{yeh2019fidelity} proposed a theoretical framework which justifies that an extension of SmoothGrad using kernel functions could improve the faithfulness of explanations as well. Agarwal \etal \cite{agarwal2021towards} prove that SmoothGrad and a variant of LIME converge to the same explanation in expectation. Our work targets post hoc improvement of stability. While previous works only proposed usable heuristics, our paper takes it further and first makes statistical tests possible for feature attribution.

\vspace{-2mm}
\section{Median Test for Feature Attribution}
In this section, we first define some related concepts. Then we introduce the one-sided MeTFA and two-sided MeTFA, respectively. All proofs are included in Appendix \ref{sec:proof} due to the space limitation.  
\vspace{-1mm}
\subsection{Overview}\label{sec:overview}

MeTFA can be applied to any algorithm that explains the prediction by evaluating feature importance. Let the prediction function be $F: (X_i)_{i\in S} \rightarrow O$, where $S$ is the input feature set, $X_i$ is an individual feature and $O$ is the model's output. Then an explanation algorithm which assigns features with importance can be denoted as $E:(F, (X_i)_{i\in S}) \rightarrow [0,1]^{|S|}$, where $|S|$ is the number of features included in $S$. For example, in image classification, $S$ is the set of all pixels, and $|S|$ equals to $w\times h$, where $w$ is the width, and $h$ is the height of the image.

Following Smilkov \etal \cite{smilkov2017smoothgrad}, MeTFA is built upon the axiom that the generated explanation should be similar if the input and the output of the prediction function are similar. Formally, let $\mathbb{P}$ be the noise set under which we presume the prediction function is robust, \ie, $O$ changes little for inputs after adding some noises sampled from $\mathbb{P}$. Similar to Smilkov \etal \cite{smilkov2017smoothgrad} and Yeh \etal \cite{yeh2019fidelity}, the first step of MeTFA is to sample noises from $\mathbb{P}$, add them to the original input, pass these noisy inputs all through the explanation algorithm and get explanations $\{e_i\},\ i=1,\dots N.$, which subject to some unknown distribution $\mathbb{D}$. We call $\{e_i\}$ the sampled explanations. $e_{ij}$ is the sampled feature score in $e_i$ for feature $j$.

% We need to define what we mean when talking about underlying saliency values. It can be defined in many ways, such as the mean of $\mathbb{D}$. However, since nothing but samples about $\mathbb{D}$ is known, we use the median of $\mathbb{D}$ which only requires to compute the cumulative density function (CDF) of $\mathbb{D}$ as the underlying saliency value because CDF is easier and more robust to estimate than the probability density function (PDF). Formally, the underlying saliency value $V = \lim_{\epsilon\rightarrow 0} (\sup \{P(x\le V)=0.5-\epsilon\} + \inf \{P(x\le V)=0.5+\epsilon\})\div 2$, where $x \sim \mathbb{D}$. In the usual case of continuous and most discrete $\mathbb{D}$, the formula can be simplified as $V = \inf \{P(x\le V)=0.5\}$.

We tackle the unknown distribution by converting $\mathbb{D}$ on the explanations to a Bernoulli distribution on a specific statistics called counting variable, based on the following two key properties of the median $V$:
\begin{itemize}
    \item If most of the samples are greater than a fixed value $h$, then $V$ is more likely to be greater than $h$, vice versa.
    \item For any continuous distribution $\mathbb{D}$ and $x\sim \mathbb{D}$, we have $P(x\ge V)=0.5$ and $P(x\le V)=0.5$. Therefore, for any $h\ge V$, $P(x\ge h)\le 0.5$, vice versa.
\end{itemize}
Based on the first property, we define the \emph{counting variable} $ct_j(h)=\sum_{i=1}^N I(e_{ij}\ge h)$, where $I$ is the indicator function. Note that $I(e_{ij} \ge h)$ follows a Bernoulli distribution with parameter $q_j(h)=P(e_{ij}\ge h)$. In addition, $e_{ij}$, $i=1,2,...,N$, is independent for any fixed $j$. Therefore, $ct_j(h)\sim B(N, q_j(h))$, where $B$ is a Binomial distribution. Then we use $ct_j$ as the test statistic to design hypothesis testing.

The second property provides a way to estimate the p-value for continuous explanation.
In practice, $\mathbb{D}$ may be discrete for some explanation methods, e.g., LIME. However, we can easily approximate any discrete distribution by a continuous distribution to any precision. Specifically, we add a very small continuous noise, \eg $\mathcal{N}(0, 10^{-6})$, to $\mathbb{D}$. For example, consider LIME which gives 0-1 map of the pixels as the explanation. The distribution of LIME's explanation on noisy inputs is discrete, as it can only be 0 or 1. Further, assume that given a particular input and a particular pixel, the probability of being 0 is 0.6, and thus the median of the distribution of LIME's explanation is 0. Then, the probability of sampling a value greater than 0 is only 0.4, not the 0.5 that we utilize. In this case, MeTFA has an assumption violation. However, if we add a small noise to LIME's explanation, say $\mathcal{N}(0, 10^{-6})$, then the LIME's explanation becomes a sharp bi-modal distribution which is concentrated around 0 and 1. The modified distribution is continuous, and thus the assumption that the probability of a sample greater than the median equals to 0.5 holds, which makes MeTFA applicable. In addition, as long as the modification noise is continuous and very small, this approximation only possibly changes the median by a tiny difference, thus the result is not affected. In particular, applying $\mathcal{N}(0, 10^{-6})$ or $\mathcal{N}(0, 10^{-8})$ does not make a difference for LIME. Note that we add small perturbation to the sampled explanations here rather than to the input, and this only aims at making the distribution of $\{e_i\}$ continuous. Therefore, for simplicity, we assume $\mathbb{D}$ is continuous in the paper.

% As we described before, selecting the median of $\mathbb{D}$ as the stable explanation can reduce the effect of abnormal extreme values. At the same time, by assuming $e_i$ is greater or smaller than a real number $h$, we can convert $\mathbb{D}$ to a Bernoulli distribution, where the single-experiment success rate $p$ can be calculated by establishing a hypothesis on the median. In this way, we can conduct hypothesis testing based on the Bernoulli distribution to establish theoretical guarantee.
In the following, we describe two different processing of $\{e_i\}$ to perform the importance test (whether $V$ is greater or smaller than $h$) in Section \ref{sec:one-side-MeTFA} and bound test (the interval that $V$ most likely falls into) in Section \ref{sec:two-side-MeTFA}, respectively. 

\vspace{-1mm}
\subsection{One-sided MeTFA}
\label{sec:one-side-MeTFA}
One-sided MeTFA is to test $H_0: V<h$ or $H_0: V>h$ for some fixed $h$. This can be broken down into two steps. First, convert $\mathbb{D}$ to a Bernoulli distribution. Second, conduct the test based on the Bernoulli distribution. Thus, in the following, we first derive how to construct the Bernoulli distribution and then conduct the statistics test. 

% Let $e_{ij}$ be the attribution score assigned by an explanation algorithm to feature $j$ in the sampled $e_i$ and $q_j(h)=P(e_{ij}\ge h)$. We further define the \emph{counting variable} $ct_j(h)=\sum_{i=1}^N I(e_{ij}\ge h)$, where $I$ is the indicator function. Since each $I(e_{ij}\ge h)$ subjects to a Bernoulli distribution with parameter $q_j(h)$, $ct_j(h)\sim B(N, q_j(h))$, where $B$ is a binomial distribution, and thus we can take $ct_j(h)$ as a test statistic. 
We have derived in Section \ref{sec:overview} that the counting variable $ct_j(h) \sim B(N, q_j(h))$, where $q_j(h)=P(e_{ij}\ge h)$.
Using this property, we are able to derive Proposition \ref{prop:one-sided-test}.

\begin{proposition}
    \label{prop:one-sided-test}
    Suppose $V_j$ is the median of $\mathbb{D}$ for feature $j$. Then, $P(ct_j(h)\ge m) \le \sum_{i=m}^N \binom{N}{i} \times p_*^i \times (1-p_*)^{N-i}$ for $V_j \le h$, where $p_*=\min(0.5, i/N)$, and $N$ is the number of sampled explanations. Similarly, $P(ct_j(h)\le m) \le \sum_{i=0}^m \binom{N}{i} \times p_*^i \times (1-p_*)^{N-i}$ for $V_j \ge h$, where $p_*=\max(0.5, i/N)$. The proof is in Appendix \ref{prof:prop:one-sided-test}.
\end{proposition}

Using Proposition \ref{prop:one-sided-test}, we can get Theorem \ref{th:one-sided-MeTFA} which enables us to perform the one-sided MeTFA.
\begin{theorem}
    \label{th:one-sided-MeTFA}
    Suppose we observe $ct_j(h)=k$. Then, the p-value of $H_0: V_j \le h$ is $\sum_{i=k}^N \binom{N}{i} \times p_*^i \times (1-p_*)^{N-i}$, where $p_*=\min(0.5, i/N)$. Similarly, the p-value of $H_0: V_j \ge h$ is $\sum_{i=0}^k \binom{N}{i} \times p_*^i \times (1-p_*)^{N-i}$, where $p_*=\max(0.5, i/N)$. The proof is in Appendix \ref{prof:th:one-sided-MeTFA}.
\end{theorem}

Therefore, to test whether $V_j$ is greater or smaller than $h$, we first count how many times $e_j$ is greater than $h$, compute the $p$-values according to Theorem \ref{th:one-sided-MeTFA}, and then compare the $p$-values with user-interested confidence level $\alpha$. From the procedure described above, we can see that the total complexity is $O(N)$ for the one-sided MeTFA.
\vspace{-1mm}
\subsection{Two-sided MeTFA}
\label{sec:two-side-MeTFA}
Two-sided MeTFA is to test $H_0: V=h$ for some fixed $h$. Under $H_0$, we have $q_j(h)=0.5$ and a too large or too small $ct_j(h)$ is rare. Using this property, we are able to derive Proposition \ref{prop:two-sided-test}.

\begin{proposition}
    \label{prop:two-sided-test}
    $P(ct_j(h)\le m_1)+P(ct_j(h)\ge m_2) \le\\ \sum_{i\in \{0,\dots,m_1\} \cup \{m_2,\dots,N\}} 0.5^N \times \binom{N}{i}$ for $V_j=h$. The proof is in Appendix \ref{prof:prop:two-sided-test}.
\end{proposition}

Proposition \ref{prop:two-sided-test} directly suggests the two-sided MeTFA, as shown in Theorem \ref{th:two-sided-MeTFA}.
\begin{theorem}
    \label{th:two-sided-MeTFA}
    Suppose we observe $ct_j(h)=k$. Let $k_1=\min(k, N-k)$ and $k_2=\max(k,N-k)$. Then, the p-value of $H_0: V=h$ is $\sum_{i\in \{0,\dots,k_1\} \cup \{k_2,\dots,N\}} 0.5^N \times \binom{N}{i}$. The proof is in Appendix \ref{prof:th:two-sided-MeTFA}.
\end{theorem}

A direct application of Proposition \ref{prop:two-sided-test} gives the confidence interval of $V$ as well. It is shown in Theorem \ref{th:interval}.

\begin{theorem}
    \label{th:interval}
    Let $k_{1} = \argmax_k \{ \sum_{i=0}^{k} 0.5^N \times \binom{N}{i} \le \frac{\alpha}{2}\}$ and $k_{2}=N-k_{1}$. Let $h_{1j}$ be the $k_{1}$-th  smallest in $\{e_{ij} \mid i=1,\dots,N\}$ and $h_{2j}$ be the $k_{2}$-th smallest. Then $(h_{1j}, h_{2j})$ is a $1-\alpha$ confidence interval for $V_j$. The proof is in Appendix \ref{prof:th:interval}.
\end{theorem}

Therefore, to test whether $V_j$ is equal to $h$, we first count how many times $e_j$ are greater than $h$, compute the $p$-values according to Theorem \ref{th:two-sided-MeTFA}, and then compare the $p$-values with custom confidence levels. To obtain the confidence intervals, we need to find the maximum $k_1$ that makes $\sum_{i=0}^{k_1} 0.5^N \times \binom{N}{i}$ smaller than $\alpha/2$, compute $k_2$ by $N-k_1$, and then get the interval from the sorted $\{e_{ij}\}$. We name the map consisting of $h_{1j}$ the \emph{lower bound map} and the map consisting of $h_{2j}$ the \emph{upper bound map}.  From the procedure described above, we can see that the complexity is $O(N)$ for the two-sided MeTFA and $O(N\log N)$ for computing the confidence interval. The pure test takes a very short time compared to the sampling process, which is bottlenecked by the speed of the explanation algorithm. However, the sampling process can be fully parallelized to take a constant time.
\vspace{-2mm}
\section{MeTFA-based Attribution Map}

In this section, based on MeTFA, we first design two kinds of maps, named \emph{MeTFA-significant map} and \emph{MeTFA-smoothed map}, to point out the significant important (unimportant) supportive features and quantify the stability of explanation, respectively. Then, we give the lower bound of the number of samples to achieve a user-interested confidence level $\alpha$.
% To point out the significant important (unimportant) supportive feature, in \ref{sec:Significant}, we design MeTFA-significant map, which uses one-sided MeTFA and clustering to visually show the significance of every feature. 
% To estimate the stability of the explanation, in \ref{sec:stability measure}, we design $\alpha$-stability, which uses the difference between the upper bound and the lower bound calculated from two-sided MeTFA to measure the stability of the explanation. 
% To increase the stability of the explanation, in \ref{sec:smoothed}, we design MeTFA-smoothed map, which uses two-sided MeTFA to calculate the upper bound and the lower bound and average all the feature score between these two bounds as the approximation of the median.

\begin{figure}[t]
    \centering
    \includegraphics[width=0.9\columnwidth]{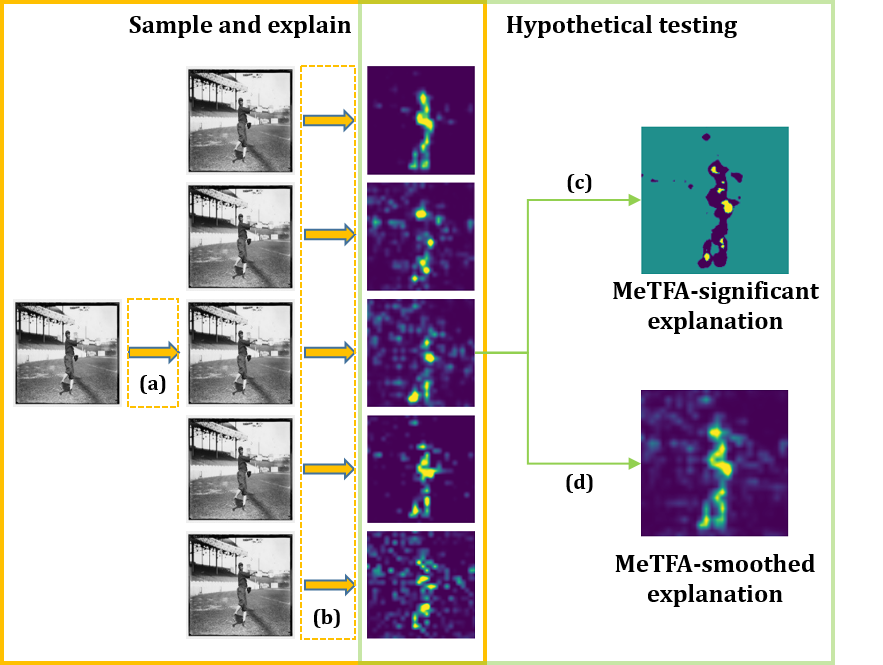}
    \caption{The overview of MeTFA. First, in (a), we add some noise from a specific distribution $\mathbb{P}$  (\eg Normal, Uniform) to the clean image. Then, in (b), we generate an explanation for each noisy image using an existing attribution method (\eg RISE, IGOS). Finally, we use the one-sided MeTFA (in (c)) to generate the MeTFA-significant map and use the two-sided MeTFA (in (d)) to generate the MeTFA-smoothed map.}
    \label{overview}
\end{figure}

\vspace{-1mm}
\subsection{MeTFA-Significance Map}\label{sec:Significant}
While the attribution maps that show every feature's importance are informative, in many cases, we only want to know what features are important and what features are unimportant. For example, when we explain the prediction of an image classification model to laypersons, they only want a subregion of the input image highlighting the important features, which motivates us to develop \emph{MeTFA-Significance Maps} to highlight the important features and unimportant features. We use the task of image classification to show the core idea.

Formally, when trying to figure out the ``important'' and ``unimportant'' features, we are actually classifying these features into two groups. Therefore, we first do a global one-dimensional two-group clustering using \emph{Jenks natural breaks algorithm} \cite{Jenks1967TheDM} for the attribution values in the sampled $\{e_{ij}\}$. This allows us to find the optimal $h$ for the two groups, \ie, $h$ is recommended to be the break threshold of these two groups. Since the ``best" $h$ varies from image to image, compared with previous works, which set a fixed threshold $h$ \cite{doan2020februus} based on their requirements, our clustering approach can select a better $h$ adaptive to the image. Next, we perform one-sided MeTFA for every feature with regard to the threshold $h$. The features whose scores are significantly greater than $h$ are classified as important, and significantly smaller than $h$ are classified as unimportant. Others are in the between, denoted as ``undecided'', meaning that they are neither significantly greater than $h$ nor significantly smaller than $h$. In the case of image classification, we paint the important features in yellow, the unimportant features in dark green, and the undecided features in dark purple. An example is presented in Figure \ref{overview}.

% \subsection{Stability Measure}\label{sec:stability measure}

% Measuring the stability of explanations is important to evaluate explanation algorithms. Many statistics, such as variance and sensitivity \cite{DBLP:conf/nips/YehHSIR19}, can be used to measure stability. These metrics are intuitive but lack specification of uncertainty involved when computing using Monte-Carlo sampling. Therefore, we propose a new metric called \emph{$\alpha$-stability} based on the two-sided MeTFA.

% \begin{definition}
%     \label{def:stability-measure}
%     Suppose the $1-\alpha$ confidence interval computed via Theorem \ref{th:interval} is $[H_1, H_2]$, where $H_i = (h_{ij})_{j=1}^{\#S}$ is the vector consisting of $h_{ij}$, $i=1,2$. $\alpha$-stability is defined to be the mean of $1-(H_2-H_1)$.
% \end{definition}

% $\alpha$-stability has several important properties. First, when the explanation is absolutely stable, \ie, $e_{i}$ are all equal, then the $\alpha$-stability is one. Second, when the explanation has the worst stability, the confidence intervals are all $[0,1]$, then the $\alpha$-stability is zero. For normal cases, it is a number in $(0,1)$. Third, the uncertainty in $\alpha$-stability computation comes from sampling explanations, quantified by $\alpha$. A more confidently computed $\alpha$-stability requires a smaller $\alpha$, and thus represents a trade-off between better-computed metrics and upper stability values. A smaller $\alpha$ is similar to a more strict judge. It scores lower for algorithms but is more confident at its scoring.

\vspace{-1mm}
\subsection{MeTFA-smoothed Map}\label{sec:smoothed}

Two-sided MeTFA can deduct a stabilization method to stabilize explanation algorithms. In the following, we first introduce the stabilized explanation named MeTFA-smoothed map, and then, we theoretically prove that the variance of theMeTFA-smoothed explanation shrinks to 0 with a same or faster speed than the SmoothGrad explanation.
% and then analyze its possessed properties, including: (1) it is unbiased, (2) it converges to $V$ at a rate of at least $O(1/N)$ and (3) it exceeds SmoothGrad \cite{DBLP:journals/corr/SmilkovTKVW17} in stability in experiments, especially for small $N$. We name this method as \emph{MeTFA-smoothed explanation}.

The MeTFA-smoothed explanation uses the mean of the explanations only included in the confidence interval rather than all of the sampled explanations. The formal definition is shown in Definition \ref{def:MeTFA-able}.

\begin{definition}
    \label{def:MeTFA-able}
    Suppose we have the sampled explanations $\{e_i\}$, and we have computed $k_1$ and $k_2$, respectively. Let $e_{(a)j}$ be the $a$-th smallest element in $\{e_{ij}\}$. Then the MeTFA-smoothed explanation is calculated as follows: the attribution score of feature $j$ is defined to be $\sum_{a=k_{1}+1}^{k_{2}-1} e_{(a)j} / (k_2-k_1-1)$. We denote it as $\ST_j$.
\end{definition}

The MeTFA-smoothed explanation has two important properties. They are summarized in Theorem \ref{th:MeTFA-able}.

\begin{theorem}
    \label{th:MeTFA-able}
    $\ST_j$ possesses the following properties. The proof is in Appendix \ref{prof:th:MeTFA-able}:
    \begin{enumerate}
        % \item $\ST_j$ is an unbiased estimation for $V_j$.
        \item $\ST_j$ converges to $V_j$ when $N$ is sufficiently large.
        \item Under the mild assumption that $f_e(V_j)>0$, where $f_e$ is the PDF of $e$, $\Var(S_j)$ converges to zero with a speed of at least $O(1/N)$.
    \end{enumerate}
\end{theorem}

\vspace{-1mm}

Theorem \ref{th:MeTFA-able} tells us that the MeTFA-smoothed explanation is a consistent estimator for $V_j$, the median of the distribution of the sampled explanations. In addition, the variance of the MeTFA-smoothed explanation shrinks to $0$, which suggests the correctness of MeTFA. Moreover, the variance shrinks with a same or faster speed than SmoothGrad, which suggests that MeTFA-smoothed explanations are more stable and efficient.  In addition, to visualize the uncertainty in the explanations, we define the upper bound map and lower bound map to be the map visualizing the corresponding upper bound and lower bound. By comparing these two bounds, one can easily find the most uncertain features and locate the almost certain features.

We provide Algorithm \ref{alg:MeTFA maps} in the Appendix to demonstrate how to compute the MeTFA-significant explanation, MeTFA-smoothed explanation, upper bound map and lower bound map in more details.

\vspace{-1mm}

\subsection{Number of Sampled Explanations}\label{sec: compute n}
The number of sampled explanations $N$ is a critical hyperparameter for MeTFA. In fact, $N$ is closely related to the confidence level $\alpha$ of the demand. For one-sided MeTFA, as shown in Theorem \ref{th:one-sided-MeTFA}, for a fixed $N$, $p\ge\binom{N}{0}\times0.5^0\times0.5^N=0.5^N$. In order to reject $H_0$, $\alpha\ge p\ge0.5^N$. Thus, the lower bound of $N$ to achieve a given $\alpha$ is $\lceil -\log_2\alpha \rceil$. Similarly, for two-sided MeTFA, as shown in Theorem \ref{th:two-sided-MeTFA}, for fixed $N$, $p\ge\binom{N}{0}\times0.5^N\times2=0.5^{N-1}$. In order to reject $H_0$, $\alpha\ge p\ge0.5^{N-1}$. Thus, the lower bound of $N$ to achieve a given $\alpha$ is $\lceil -\log_2\alpha \rceil+1$. However, these are the minimum number of samples required, and we recommend to use more samples whenever the computational cost of the sampling explanations is acceptable.

\vspace{-2mm}
\section{Experiments}

In this section, we first explain the experiment settings in details. Then, we demonstrate the quality of MeTFA explanations from three perspectives: visualization, stability and faithfulness. 
% Then, to quantitatively guide the selection of the number of samples for sample-based methods, we propose a metric named \emph{overall confidence}.  
Finally, we discuss the impact of important hyperparameters for MeTFA.
\vspace{-1mm}
\subsection{Settings}\label{sec:setting}

We evaluate MeTFA on the image classification and the text classification task because most feature attribution methods target these two tasks. The evaluation settings is as follows:

\vspace{-1mm}
\subsubsection{Datasets and Models}\label{sec:dataset and model}
% \newline
For the image classification task, we use ILSVRC2012 validation set \cite{ILSVRC15} as the source dataset, because it is the most evaluated dataset among explanation methods. We use the pre-trained VGG16, Resnet50 and Densenet169 from PyTorch \cite{NEURIPS2019_9015} as the models to be explained. 

For the text classification task, we choose the dataset from Toxic Comment Classification Challenge\footnote{https://www.kaggle.com/c/jigsaw-toxic-comment-classification-challenge/data} which contains 159,571 training texts and 63,978 testing texts with six toxic comment classes including toxic, severe toxic, obscene, threat, insult and identity hate. We train a bidirectional LSTM concatenated with two fully connected layers on the training set, which achieves an accuracy rate of 97.46\%. Since a sentence consisting of too many words will greatly increase the number of samples required by LENMA, in the following experiment, we only use the sentences of length between $40$ and $80$. We extensively evaluate MeTFA on the IMDb Movie Reviews dataset \cite{maas-EtAl:2011:ACL-HLT2011}. We split the $50000$ reviews into training set (containing $40000$ reviews) and test set (containing $10000$ reviews), and the same model is applied. The trained model achieves an accuracy rate of $88.39\%$ on the test set.
\vspace{-1mm}
\subsubsection{Explanation Algorithms}\label{sec:source explanations}

In the image domain, we apply MeTFA to four types of mainstream feature attribution methods, including Gradient (the most classic gradient-based method), RISE and LIME (two most popular sample-based methods), IGOS (the SOTA optimization-based method) and ScoreCAM (the SOTA CAM-based method).  In the  text domain, we apply MeTFA to LENMA, the SOTA explanation method designed for RNN.

% In text classification task, we choose the dataset from Toxic Comment Classification Challenge\TBD{cite} which contains 159571 training texts and 63978 testing texts with six toxic comment classes including toxic, severe toxic, obscene, threat, insult and identity hate. We train a bidirectional LSTM concatenated with two full connected layers on the training set, which achieves an accuracy rate of 97.46\%. We apply MeTFA to LENMA, the SOTA explanation method designed for RNN.
\vspace{-1mm}
\subsubsection{Metrics}  \label{sec:evaluate_matric}
We introduce the metrics to evaluate the stability of explanations and the faithfulness of explanations

\paragraphbe{Stability} 
The standard deviation is a good choice to measure the variety of the output. 
To measure the stability of a feature attribution method under noises, we use the \emph{mstd} (short for the mean of \emph{std}) metric, defined to be the mean of the standard deviation of explanations on noisy inputs sampled from the neighborhood of the original input.
The formal definition of mtsd is as follows:
$$
\text{mstd}=Mean_{I\in D, i\in S}(std_{n\sim\mathbb{O}_n}(E(I+n)_i))
$$
where $E(\cdot)_i$ returns the attribution score of feature $X_i$ and $D$ is the test data set. 
% To better simulate the real-world noise, mstd is calculated under different noise distribution $\mathbb{O}_n$, e.g., Normal, Uniform, etc. 
The default number of noisy inputs is set to 10, \ie, for every image, we sample $10$ noises $n$ from $\mathbb{O}_n$. By the definition, a lower mstd value means a more stable explanation.

%We propose \emph{mstd} (\emph{std} for the explanation) based on the most popular metric standard deviation (\emph{std}) to evaluate the stability of explanations in the both image and text domain. Formally, supposing the distribution of outer noise is $\mathbb{O}_n$ and for an input $I = (X_i)_{i\in S}$, an explanation algorithm $E$ attributes a contribution score for every feature $X_i$. Then \emph{std} is defined as following.
%\[mstd=Mean_{i\in S}(std_{n\sim\mathbb{O}_n}(E(I+n)_i))\]
%where $E(\cdot)_i$ returns the attribution score of feature $X_i$.
% In the text domain, we use two metrics to evaluate the stability: \emph{std} and \emph{overlap of top n}. Similar to the image domain, we use \emph{std} as a metric. Moreover, we compute the overlap of the $n$ words with top feature scores when adding different noises, since users may just consider about the top $n$ important words.

\paragraphbe{Faithfulness} The faithfulness metric is used to measure whether the features highlighted by an attribution map support a model's prediction. An explanation is called faithful if the generated attribution maps highlight the supportive features. In our experiments, we use two kinds of metrics to evaluate the faithfulness. One is the most popular metric used in the image and text domain proposed in the previous work, and the other one is our proposed more robust faithfulness metric. 
% As defined in most papers, When an explanation is faithful to a model, then the explanation locates the supportive feature accurately. 

In the image domain, \emph{insertion}, \emph{deletion} \cite{petsiuk2018rise} and \emph{overall} \cite{zhang2021group} are commonly used to estimate the faithfulness of an attribution map. These metrics aim to measure whether the features highlighted by an attribution map support a model’s prediction. If an attribution map is faithful to the model, then removing the pixels with the highest values from a full image will cause a big decrease on the predicted score and conversely, inserting the pixels with the highest values into a blank image will cause a big increase on the predicted score. To formally illustrate this property, let $I$ be the original image, $f_c(\cdot)$ returns the predicted score of label $c$ and $M$ is the attribution map given by a explanation algorithm for $I$. Then, the (normalized) insertion and deletion are defined as follows:
\[ insertion(I, M) = \frac{1}{f_c(I)}\int_0^{100}f_c(I_n^{M+})dn\]
\[ deletion(I, M) = \frac{1}{f_c(I)}\int_0^{100}f_c(I_n^{M-})dn\]
where $I_n^{M+}$ keeps the $n\%$ pixels in $I$ with top attribution scores, and $I_n^{M-}$ deletes the $n\%$ pixels in $I$ with top attribution scores. A more faithful attribution map can highlight the supportive features more accurately and thus keeping the same number of pixels can get a higher predicted score, resulting in a higher insertion value. Similarly, a more faithful attribution map gets a lower deletion value. To specifically show the process, we add an example in the appendix (Figure \ref{fig:insertion and deletion}). Besides, sometimes insertion and deletion give the contradictory results. In such situation, overall, which is equal to insertion minus deletion, is used to evaluate the faithfulness. However, the faithfulness metrics considering only a single point suffers from the effect of non-robust features. Therefore, we take the neighborhood of the input into consideration. Specifically, we propose \emph{robust insertion} (RI), \emph{robust deletion} (RD) and \emph{robust overall} (RO) to further evaluate the faithfulness for the neighborhood of the clean image, which are as follows.
\[ RI(M, I, \mathbb{O}_n) = \mathbb{E}_{n\sim\mathbb{O}_n}(insertion(I+n,M))\]
\[ RI(M, I, \mathbb{O}_n) = \mathbb{E}_{n\sim\mathbb{O}_n}(insertion(I+n,M))\]
where $\mathbb{O}_n$ is the distribution of noise. 
% The calculation of RI and RD is as follows. First, we generate an attribution map with an original image, naming it an original map. Then, we add different noises to the original image, obtaining some noisy images. Finally, RI (RD) is calculated as the average insertion (deletion) of the original map and the noisy images. 
The intuition behind the RI (RD) metric is that if the attribution map finds the robust supportive features, then after adding a small random noise, the features will still keep supporting and thus having a high insertion score or low deletion score on the noisy images. Therefore, a higher robust faithfulness means the explanation can locate the robust features more precisely. In practice, we use $10$ samples to approximate the expectation. Correspondingly, RO is defined as RI minus RD. By the definition, a lower value of deletion or RD suggests a more faithful explanation while a higher value of insertion, overall, RI or RO suggests a more faithful explanation.

In the text domain, we use \emph{Feature Deduction Test} (FDT), \emph{Feature Augmentation Test} (FAT) and \emph{Synthetic Test} (ST) proposed in LEMNA to estimate the faithfulness of an explanation. Similar to the image task, let $T$ be the original text, $f_c(\cdot)$ returns the predicted score of label $c$ and $M$ is the attribution map given by a explanation algorithm for $T$. Then FDT, FAT, ST can be defined as follows:
\[FDT(T,M) = f_c(T^{M-}_n)\]
\[FAT(T,M,T') = f_c(T'\circ T_n^{M+})\]
\[ST(T,M) = f_c(T^{M+}_n)\]
where $T^{M-}_n$ deletes the $n$ words in $T$ with top attribution scores,  $T'\circ T_n^{M+}$ retains the $n$ words with top attribution scores in $T$ but replaces the other words by a randomly selected instance $T'$, and $T^{M+}_n$ only keeps the $n$ words in $T$ with top attribution scores. 
% A more faithful attribution map highlights the supportive features more accurately, thus deleting the same number of features can get a lower predicted score which causes a lower FDT value. 
% The specific implementation is as follows. FDT sets the values of the $n$ features with top attribution scores to $0$ and then tests the predicted score of the modified input. Thus a more faithful explanation will get a lower predicted score. To test the faithfulness of an explanation to an original instance, FAT firstly selects a random instance from the source dataset, and then transplant the $n$ features with top attribution scores in the original instance to the selected instance and then tests the predicted score of the transplanted instance on the original instance's predicted label. Thus a more faithful explanation will get a higher predicted score. ST keeps the $n$ features with top attribution scores and sets the other features to $0$. Thus a more faithful explanation will get a higher predicted score for the modified input. 
Similar to the image domain, we extend these three metrics to the robust faithfulness metrics, \ie, RFDT, RFAT and RST, and use 10 samples for calculation. By the definition, a lower value of FDT or RFDT suggests a more faithful explanation while a higher value of FAT, ST, RFAT or RST suggests a more faithful explanation.
% For every test, we first generate an explanation of the original text. Then we add noises to the text and test the faithfulness of the original explanation on the noisy texts. Finally, the value of the robust faithfulness metric is equal to the average of the original explanation on the noisy texts.
\[RFDT(T,M,\mathbb{O}_n) = \mathbb{E}_{n\sim\mathbb{O}_n}(FDT(T+n,M))\]
\[RFAT(T,M,\mathbb{O}_n,T') = \mathbb{E}_{n\sim\mathbb{O}_n}(FAT(T+n,M,T'))\]
\[RST(T,M,\mathbb{O}_n) = \mathbb{E}_{n\sim\mathbb{O}_n}(ST(T+n,M))\]
% In image classification, the metrics are selected as follows. For stability, we use the most popular metric standard deviation (\emph{std}) to evaluate the stability of explanations. For faithfulness, we use the most popular metrics \emph{insertion} and \emph{deletion} to evaluate the ability of locating the supportive features. To evaluate the robustness of support features to random noise, we propose two metrics \emph{robust insertion} and \emph{robust deletion}. 

% In text classification, the metrics are selected as follows. For stability, similar to the image domain, we use the most popular metric standard deviation (\emph{std}) to evaluate the stability of explanations. For faithfulness, we use the metrics proposed with LEMNA, \ie, \emph{Feature Deduction Test}, \emph{Feature Augmentation Test} and \emph{Synthetic Test}. To evaluate the robustness of support features to random noise, similar to the image domain, we propose three corresponding metrics.

% Sampling number is a critical hyperparameter of sample-based methods. To quantitatively select this hyperparameter, we design a metric named \emph{overall confidence} based on the two-sided MeTFA to quantitatively guide hyperparameter selection.

The roles of the noise distribution $\mathbb{P}$ and $\mathbb{O}_n$ are different. $\mathbb{P}$ is used to sample around the original data, which is a core step of MeTFA, while $\mathbb{O}_n$ is used to compute the robust metrics. We take RI as an example to show the difference more specifically and we provide its algorithm in Algorithm \ref{alg:calculation RI} in the appendix. 
\vspace{-1mm}
\subsubsection{Default Settings}
Unless otherwise specified, all the hyperparameters are set as follows. 
\begin{itemize}
    \item The confidence level $\alpha$. We set a common choice $\alpha=0.05$. 
    \item The number of sampled explanations $N$. As discussed in Section \ref{sec: compute n}, the minimum $N$ to achieve $\alpha = 0.05$ for MeTFA-significant map is $5$. However, this leads to too few features being tested significant. Therefore, we choose $N=10$ as the default.
    \item The number of samples for the sample-based methods. Typically, the more samples used by the sample-based method, the more stable the generated explanation would be. For LIME, we choose $1000$ because this is the default in the LIME's open-source code. For RISE, we choose $1000$ as well so that it is consistent to LIME since they are compared to each other in the image domain.  For LEMNA,  we use 500 and 2000 for the Toxic Comments dataset and 2000 for the IMDb Reviews dataset.
    \item For the other parameters, we use the same as the corresponding open-source code.
\end{itemize}

\begin{figure}
     \begin{subfigure}[c]{0.15\columnwidth}
         \centering
         \includegraphics[width=\columnwidth]{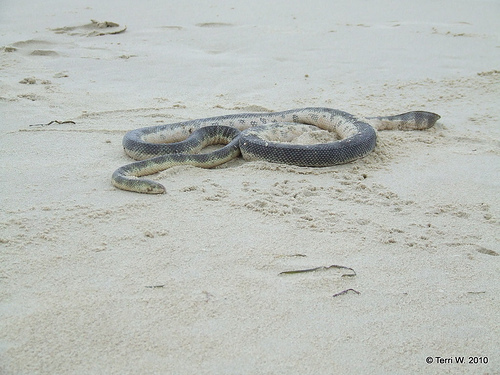}
         \caption{}
         \label{image}
     \end{subfigure}
     \begin{subfigure}[c]{.8\columnwidth}
         \includegraphics[width=\columnwidth]{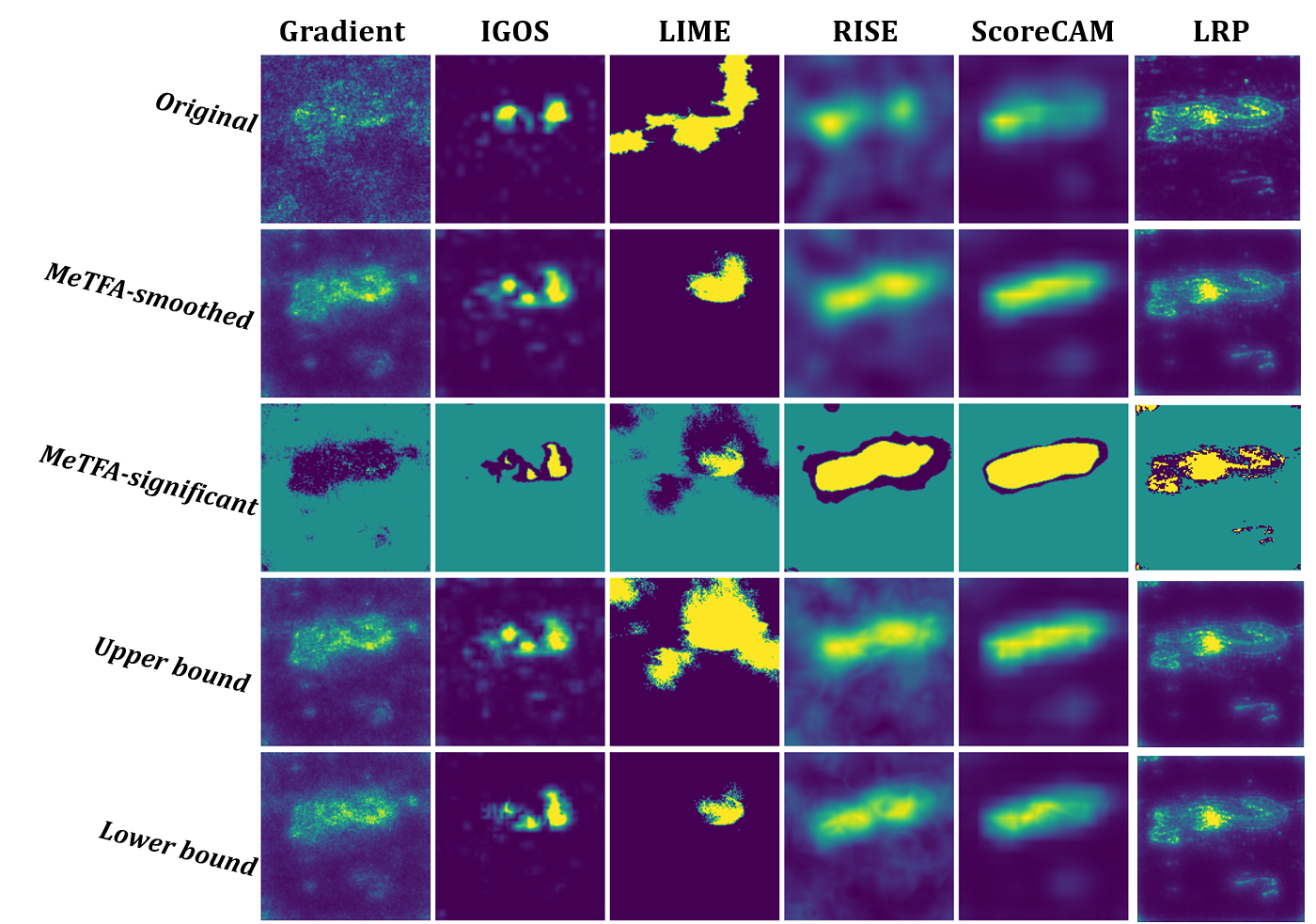}
         \caption{}
         \label{fig:maps}
     \end{subfigure}
    \caption{Visual examples for the original, the MeTFA-smoothed and the MeTFA-significant  explanations. (a) is the input image. The columns of (b) show the explanations of RISE, IGOS, LIME, Gradient and ScoreCAM, respectively.}
    \label{visualize_mask}
\end{figure}
\vspace{-1mm}
\subsection{Quality of Visualization}\label{sec:visualization}

In this part, we illustrate the visualization effect of MeTFA in the image domain. We use the pre-trained VGG16 network from the Pytorch and an image from the source dataset as an example. The predicted label of the image is ``sea snake'', which is consistent to the ground truth. We apply the five feature attribution methods, as discussed in Section \ref{sec:source explanations}, to explain this prediction. The results are shown in Figure \ref{visualize_mask}.
%Different columns in Figure \ref{fig:maps} are the $5$ different explanations algorithms and different rows are the different kinds of process of MeTFA. In the following, we demonstrate the meaning of the different maps row by row.

The original explanations, shown in the first row of Figure \ref{visualize_mask}, are roughly located around the sea snake, which is intuitive. The second and the third row contains the MeTFA-smoothed and MeTFA-significant maps, generated by applying the two-sided and the one-sided MeTFA to Gradient, IGOS, LIME, RISE and ScoreCAM, respectively. Apparently, the second and third row have better visual quality than the first row. For example, the MeTFA-smoothed Gradient highlights the snake while the original explanation is scattered, and the MeTFA-significant RISE shows that only the snake area is significantly important while the original explanation contains a lot of noises. In general, MeTFA leaves fewer pixels as ``undecided'' if the original explanation is more stable, \eg, ScoreCAM, and a less stable explanation can benefit more from applying MeTFA, \eg, Gradient. The last two rows show the upper bound maps and lower bound maps. We can directly find which explanations are more stable and which features are uncertain by comparing these two maps. For example, LIME has low stability as the lower bound map and the upper bound map are greatly different while ScoreCAM has high stability as the lower bound map.

\vspace{-1mm}
\subsection{Stability}\label{sec:stability}
In this part, we evaluate the stability of MeTFA in the image and text domain. As discussed in Section \ref{sec:overview}, MeTFA samples noises from a distribution $\mathbb{P}$ to estimate the distribution of the explanations. In addition, as discussed in Section \ref{sec:evaluate_matric}, the metric mstd measures the stability of an attribution method by sampling from a noise distribution $\mathbb{O}_n$.
% the stability metric samples noises from a distribution $\mathbb{O}_n$. 
In practice, $\mathbb{O}_n$ might be different to $\mathbb{P}$. Therefore,
to make the setting more representative for the real applications, we evaluate each combination of $\mathbb{P}$ and $\mathbb{O}$ with several common noises.% in the following.
\vspace{-1mm}
\subsubsection{Stability in the Image Domain} \label{sec:image-stability}
In the image domain, 
% to test the stability of MeTFA-smoothed explanation to various noises, we test under four common noise scenarios in real life to simulate the real world. The four kinds of noises are as follows:
$\mathbb{P}$ and $\mathbb{O}_n$ are selected from the following three noise distributions which are very common for images: (1) uniform distribution  $U(-0.1,0.1)$, (2) normal distribution $N(0,0.01)$, (3) brightness, \ie,  multiply by a factor $n$, $n\sim U(0.9,1.1)$. Although real-world noises may have joint patterns, we apply these perturbations independently to each pixel of the image to simulate the noises introduced by the image sensor \cite{wiki:Image_noise}, \eg, a camera. In addition, neural networks are empirically robust to such random noises \cite{10.5555/3157096.3157279}, thus the correct explanation is probable to remain the same under the random noises, which makes the median value suitable for explaining the original input.

\begin{table}
\centering
\caption{The mstd value of the MeTFA-smoothed RISE and vanilla RISE for Densenet169. The last column is the average mstd among three kinds of $\mathbb{O}_n$ for a fixed $\mathbb{P}$. }
\small
\label{tab:stability RISE Densenet}

\begin{tabular}{c c c c c} 
\toprule
$\mathbb{P}\backslash\mathbb{O}_n$& Normal & Uniform & Brightness & Avg\\
\midrule
Normal & \textbf{0.0591} & \textbf{0.0491} & 0.0406 & \textbf{0.0496}\\
Uniform & 0.0613 & 0.0504 & 0.0413 & 0.051\\
Brightness & 0.0668 & 0.0524 & \textbf{0.0366} & 0.0519\\
\midrule
Vanilla RISE & 0.1197 & 0.1219 & 0.1074 & 0.1163\\
\bottomrule
\end{tabular}
\end{table}

% The noise distribution in the natural environment might be unknown. In this part, we suppose the Laplace distribution is our unknown noise. That means that the sampling distribution ($\mathbb{P}$) illustrated in \ref{sec:overview} of MeTFA can be Uniform, Normal or Brightness, while the outer noise (denoted as $\mathbb{O}_n$) can be Uniform, Normal, Laplace or Brightness. Then,
We compare the stability of the MeTFA-smoothed explanation using different $\mathbb{P}$ with the vanilla explanation under different $\mathbb{O}_n$. To compute the mstd, we randomly select $100$ images as the test data set $D$. The results for Densenet169 with RISE algorithm are shown in Table \ref{tab:stability RISE Densenet}. Table \ref{tab:stability RISE Densenet} shows that every $\mathbb{P}$ can improve the stability under every $\mathbb{O}_n$ when compared to the vanilla RISE, decreasing the mtsd by roughly a half. Therefore, MeTFA does not need to know the ``correct'' $\mathbb{O}_n$, because the stability transfers across the noise distributions. Extensive experiments on other algorithms and other models are shown in Table \ref{tab:LIME_std}, Table \ref{tab:IGOS_std}, Table \ref{tab:ScoreCAM_std} and Table \ref{tab:stability Gradient Resnet} in the Appendix. The results show that MeTFA can significantly increase the stability of LIME, IGOS, Gradient and RISE but has slight effect on ScoreCAM. This may be because LIME and RISE are affected by the sampling process, while Gradient and IGOS are affected by the non-robust features. MeTFA can attenuate the effects of these two factors and thus shows significant increases in the stability. However, ScoreCAM does not involve a sampling process and suffers little from the non-robust features, as shown by a small mtsd for the vanilla explanations. Therefore, ScoreCam benefit less from MeTFA in this sense. 

% However, we should notify the readers that this is because these distributions we discussed only perturbs the noises slightly, and the perturbed input is in the neighborhood of the original input. With a large perturbation, this might not be the case.

Further, as we can see, the best choice of $\mathbb{P}$ to increase the stability varies for different $\mathbb{O}_n$. For example, when $\mathbb{O}_n$ is Normal, the best $\mathbb{P}$ is Normal; when $\mathbb{O}_n$ is Brightness, the best $\mathbb{P}$ is Brightness. Since the distribution of real-world noise is usually unknown, we take the average among the mstd under three kinds of $\mathbb{O}_n$ for every fixed $\mathbb{P}$ to comprehensively compare the ability of each $\mathbb{P}$ to improve the stability. The results are shown in the last column of Table \ref{tab:stability RISE Densenet}. As we can see, Normal is the best choice for $\mathbb{P}$ to increase the stability under various noises. 
Extensive experiments on other algorithms and models confirm this conclusion as well, as shown in Table \ref{tab:LIME_std}, Table \ref{tab:IGOS_std}, Table \ref{tab:ScoreCAM_std} and Table \ref{tab:stability Gradient Resnet} in the Appendix. Therefore, although normal distribution is not always the best choice for $\mathbb{P}$, it is a good default for applying MeTFA. 

Although we have established theoretical results that MeTFA is able to quantify the uncertainty better and converges as fast as SmoothGrad (short for SmoothGrad \cite{smilkov2017smoothgrad}), their stability under noises on the input is not compared. Therefore, we empirically compare the stability of MeTFA with SG  in two settings: (1) there is no noise on the input, which verifies our proof, and (2) there are noises on the input. As we introduced in Section \ref{sec:related work}, SG samples from the neighborhood of the original image and simply takes the average of all the sampled explanations. However, MeTFA takes the average of the explanations between the lower and upper bound computed from the two-sided MeTFA. This is the only difference between MeTFA-smoothed explanations and the SmoothGrad explanations. For a fair comparison, MeTFA and SG use the same noises sampled from $\mathbb{P}$.
Specifically, we randomly sample $100$ images from the source dataset as the test data set $D$ to compute mstd. As SG is designed to remove the noise for the Gradient, we take the Gradient as the representative explanation and then compare the stability for MeTFA-smoothed Gradient and SG Gradient.

Since both MeTFA and SG use sampling, their outputs naturally have randomness even if there is no external noise. Thus, we first apply no noise on the input to test the stability of MeTFA and SG, which should verify our proof about the stability advantage of MeTFA. $\mathbb{P}$ is set to be Uniform or Normal. Table \ref{Resnet vs_sg divide outer_none} shows  the ratios of the mstd of the MeTFA-smoothed Gradient over the SG Gradient, \ie, $mstd_{MeTFA-Grad}/mstd_{SG-Grad}$. It can be found that all the ratios are lower than $1$, which suggests that the MeTFA-smoothed explanations are more stable than the SG explanations when there is no external noise. Further experiments  on the VGG16 model confirms this conclusion, are shown in Table \ref{vgg vs_sg divide outer_none} in the appendix. However, this ratio increases when $N$ becomes larger, thus empirically suggests that our asymptotic bound on the convergence is tight, \ie, the lower bound for its convergence rate is $O(1/N)$ as well.

Then we test the stability when there are external noises. Specifically, the $\mathbb{O}_n$ and $\mathbb{P}$ are set to be the same and selected from Uniform or Normal. Similar to the noise-free case, we record the ratios of the mstd of the MeTFA-smoothed Gradient over the SG Gradient. Table~\ref{Resnet vs_sg divide} shows that all the ratios are lower than $1$, meaning that MeTFA is always more stable than SG. Moreover, in Appendix \ref{prof:th:MeTFA-able}, we prove that MeTFA only takes the average of $O(\sqrt{N})$ sampled explanations to generate the MeTFA-smoothed explanation. Therefore, the MeTFA-smoothed explanation averages far less sampled explanations than SG (which takes the average of $N$ sampled explanations) to obtain a higher stability because it automatically filters out abnormal extreme values. This property helps MeTFA to be even more stable than SG when the input is noisy.

In conclusion, MeTFA-smoothed explanations are more suitable when the vanilla explanations are vulnerable to the effect of non-robust features (e.g., Gradient, IGOS) or the sampling process (e.g., RISE, LIME).

% under the following two situations:
% \begin{itemize}
%     \item No outer noise and $\mathbb{P}$ is Uniform or Normal. No outer noise means $P(n=0)=1,\ n\sim\mathbb{O}_n$. This situation aims to compare the stability of MeTFA with SG when no outer noise.
%     \item $\mathbb{P}$ and $\mathbb{O}_n$ are both Uniform or Normal. This situation aims to compare the stability of MeTFA with SG when existing outer noise.
% \end{itemize}

% The results with different number of sampled explanations ($N$) are shown in Table \ref{Resnet vs_sg divide outer_none} for case without outer noise and Table \ref{Resnet vs_sg divide} for case with outer noise. It can be found that all the ratios of $mstd_{MeTFA-Grad}/mstd_{SG-Grad}$ are lower than $1$, which suggests that MeTFA-smoothed explanations are more stable than SG explanations. In fact, in the appendix (Section \ref{prof:th:MeTFA-able}), we prove that MeTFA only average $O(\sqrt{N})$ sampled explanations to generate MeTFA-smoothed explanation. Thus, MeTFA-smoothed explanation averages far less data than SG (which average $N$ sampled explanations) to obtain higher stability regardless of the existence of the external noises.

\begin{table}
\begin{center}
\caption{The results of $mstd_{MeTFA-Grad}/mstd_{SG-Grad}$ for Resnet50 under two settings: $\mathbb{P}$ is Normal or Uniform and no external noise.}

\label{Resnet vs_sg divide outer_none}
\small
\begin{tabular}{c c c} 
\toprule
$N$ & Uniform & Normal\\
\midrule
10 & 0.9451 & 0.9372\\
30 & 0.9576 & 0.9452\\
50 & 0.9693 & 0.9560\\
70 & 0.9807 & 0.9673\\
\bottomrule
\end{tabular}
\end{center}

\end{table}

\begin{table}
\begin{center}
\caption{The results of $mstd_{MeTFA-Grad}/mstd_{SG-Grad}$ for Resnet50 under two settings: $\mathbb{P}$ and $\mathbb{O}_n$ are both Uniform or Normal.}

\label{Resnet vs_sg divide}
\small
\begin{tabular}{c c c c} 
\toprule
$N$ & Uniform & Normal\\
\midrule
10 & 0.9484 & 0.9437\\
30 & 0.9166 & 0.9097\\
50 & 0.9059 & 0.8983\\
70 & 0.8996 & 0.8915\\
\bottomrule
\end{tabular}
\end{center}

\end{table}
\vspace{-1mm}
\subsubsection{Stability in the Text Domain} In the text domain, we use synonym substitution as noise, because different words express similar meanings in a sentence. Formally, this noise $\mathbf{P}(p)$  replaces every word by its synonym independently with probability  $p$.
In this experiment, we set both $\mathbb{P}$ and $\mathbb{O}_n$ to $\mathbf{P}(0.5)$. Specifically, we use wordnet in nltk \cite{bird2009natural} for synonym substitution and do not require the predicted class to keep the same in the experiment, as we want to simulate the noise in the real world. As discussed in Section \ref{sec:setting}, we use LEMNA as the target explanation and a bidirectional LSTM as the target model. Similar to Section \ref{sec:image-stability}, We randomly select 100 toxic texts whose number of words are between $40$ and $80$ as the test data set $D$.

The result is shown in Table \ref{std of lemna}, where the number of samples of LEMNA is 2000. It can be found that the mstd value of MeTFA-smoothed LEMNA is significantly smaller than that of the vanilla LEMNA, which means that MeTFA can increase the stability of LEMNA as well. The results for the Toxic Commnet, where the number of samples of LEMNA is 500, are shown in the appendix (Table \ref{tab:lemna vs metfa vs sg}), and the conclusion is the same, i.e., MeTFA can increase the stability of LEMNA.

\begin{table}[t]
\begin{center}
\caption{The mstd values for LEMNA and MeTFA-smoothed LEMNA when $\mathbb{P}$ and $\mathbb{O}_n$ are both $\mathbf{P}(0.5)$.}
\label{std of lemna}

\small
\begin{tabular}{c  c c} 
\toprule
Dataset &  LEMNA &  MeTFA-smoothed LEMNA\\
\midrule
Toxic Comment & 0.1803 & \textbf{0.0891}\\
IMDb Reviews & 0.3012 & \textbf{0.1691}\\
\bottomrule
\end{tabular}
\end{center}

\end{table}
\vspace{-1mm}
\subsection{Faithfulness}\label{sec:faithful}
Faithfulness to the model is an essential property for an explanation. The MeTFA-smoothed explanation approximates the median of the attribution maps under some $\mathbb{P}$. The following experiments shows that MeTFA greatly increases the robust faithfulness while maintaining the faithfulness level. Therefore, MeTFA-smoothed explanations find more robust features used by the model.
\vspace{-1mm}
\subsubsection{Faithfulness in the Image Domain}In the image domain, we use insertion, deletion, overall, RI, RD and RO to estimate the faithfulness of an explanation which are used to measure whether the features highlighted by an attribution map support a model’s prediction. However, the gradient-based explanations are not designed to highlight the supportive features, and thus these metrics are not suitable for them. Moreover, these metrics can only evaluate the continuous attribution maps while LIME generates a discrete (in fact, a binary) map. Therefore, we do not test these metrics for Gradient, LRP and LIME and only test them for IGOS, ScoreCAM and RISE.
The result is evaluated on the VGG16 model, and the metrics are averaged on 1000 randomly chosen images. 
% We use insertion and deletion to evaluate the six attribution maps for every image. When calculating the value of insertion and deletion, we every time add (or remove) $1\%$ pixels and get the predicted score to draw the probability curve. To keep the probability curve of deletion starting from $1$ and the probability curve of insertion end with $1$, we normalize the probability curve by dividing it by the predicted score of the complete image. 
% We experiment in the following two situations:
% \begin{itemize}
%     \item For every image, we calculate the insertion (deletion) score.
%     \item For every image, we add $10$ different noise $n\sim\mathbb{O}_n$. Then we use the attribution map of the clean image to calculate $10$ insertion and deletion score. We average $10$ scores to get robust insertion (deletion) score.
% \end{itemize}
%We use $(HB+LB)/2$ as the median estimate, and test their insertion and deletion values on 1000 randomly selected images from the ImageNet dataset. \ref{faithfulness} shows the comparative evaluations of state-of-the-art approaches using or not using MeTFA in terms of the deletion and insertion metrics on the ImageNet dataset with VGG16 as the baseline model. From \ref{faithfulness} we observe that MeTFA keeps the faithfulness of the base explanation and even improved ScoreCAM's score on insertion. This is also in line with our intuition. Because MeTFA estimates the median of the distribution of the saliency map, its faithfulness to model should also be similar to the base explanation.
In this experiment, $\mathbb{P}$ is set to be Uniform, and $\mathbb{O}_n$ is chosen from Uniform and Normal. 
% We experiment under Uniform and set $\mathbb{O}_n$ choose two different Noise distribution $N(0,0.1)$ (different from the inner\_distribution of MeTFA) and $U(-0.1,0.1)$ (same as the inner\_distribution of MeTFA) in situation two. 
% Besides, we find that the results of insertion and deletion are sometimes contradictory. For example, in Table \ref{tab:insertion}, the insertion of MeTFA-smoothed IGOS is 0.3881, which is greater than the insertion of IGOS (0.3360), and thus MeTFA-smoothed IGOS is more faithful to the model. However, the deletion of MeTFA-smoothed IGOS (0.1107) is also greater than the insertion of IGOS (0.1002), which suggests IGOS is better. Thus, we also use the overall metric \cite{zhang2021group} which is calculated by \emph{insertion-deletion}. In general, the higher (robust) insertion and (robust) overall is better, and the lower (robust) deletion is better. 
\begin{figure*}
    \centering
    \includegraphics[width=0.95\textwidth]{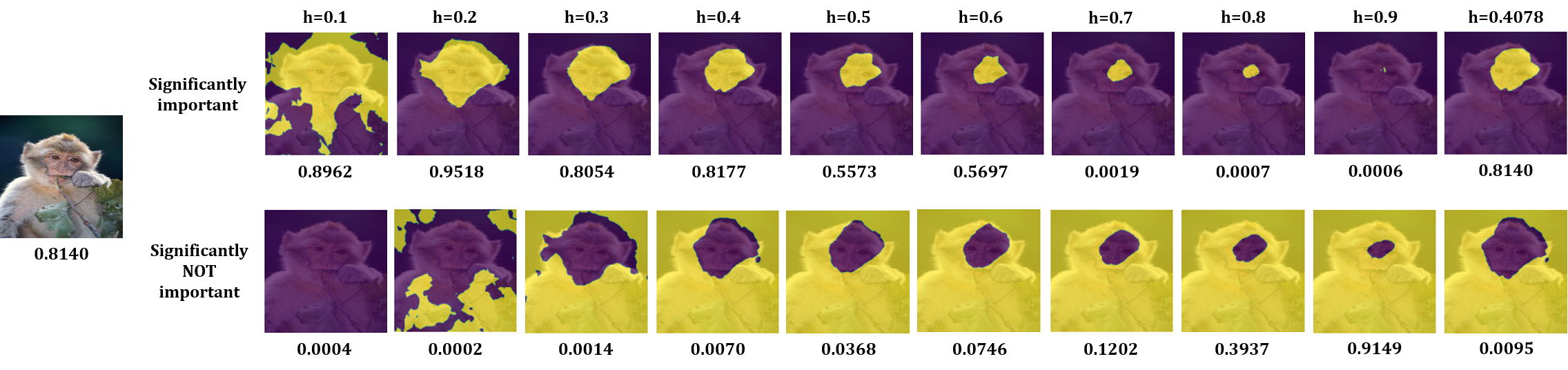}
    
    \caption{The MeTFA-significant RISE with $h$ from $0.1$ to $0.9$. The last column is the MeTFA-significant explanations with the recommended threshold $h$. The values under the images represent the predicted score of the original label, \ie, macaque, for the significantly important areas.}
    
    \label{test h}
\end{figure*}

\begin{table}
\begin{center}
\caption{The results with the faithfulness metrics. The tuple in the table is structured as (the score of the MeTFA-smoothed explanation, the score of the vanilla explanation).}

\label{tab:insertion}
\small
\begin{tabular}{c c c c} 
\toprule
method & insertion & deletion & overall\\
\midrule
ScoreCAM & (0.5897,\textbf{0.6101}) & (0.1571,\textbf{0.1439})& (0.4326,\textbf{0.4662})\\
RISE & (0.5508,\textbf{0.5556}) & (0.1767,\textbf{0.1550}) & (0.3741,\textbf{0.4006})\\
IGOS & (\textbf{0.3881},0.3360) & (0.1107,\textbf{0.1002}) & (\textbf{0.2774},0.2358)\\
\bottomrule
\end{tabular}
\end{center}

\end{table}
% \begin{table}
% \begin{center}
% \caption{The value of insertion and deletion for situation one. The tuple in the table is structured as (the score of MeTFA-smoothed explanation, the score of the original explanation)}
% \label{tab:insertion}
% \begin{tabular}{c c c c} 
% \toprule
% method & insertion & deletion & overall\\
% \midrule
% ScoreCAM & (0.590,0.610) & (0.157,0.144)& (0.433,0.466)\\
% RISE & (0.551,0.556) & (0.177,0.155) & (0.374,0.401)\\
% IGOS & (0.388,0.336) & (0.111,0.100) & (0.277,0.236)\\
% \bottomrule
% \end{tabular}
% \end{center}
% \end{table}

\begin{table}
\begin{center}
\caption{The results with the robust faithfulness metric. $\mathbb{O}_n=N(0,0.1)$. The structure of the table is similar to Table \ref{tab:insertion}.}

\label{tab:normal insertion}
\small
\begin{tabular}{c c c c} 
\toprule
method & RI & RD & RO\\
\midrule
ScoreCAM & (\textbf{1.2915},0.9429) & (0.4279,\textbf{0.4039})& (\textbf{0.8636},0.5390)\\
RISE & (\textbf{1.2959},0.8626) & (\textbf{0.4497},0.4846)& (\textbf{0.8462},0.3780) \\
IGOS & (\textbf{0.5061},0.4152) & (0.2249,\textbf{0.1640})& (\textbf{0.2812},0.2512) \\
\bottomrule
\end{tabular}
\end{center}

\end{table}
% \begin{table}
% \begin{center}
% \caption{The value of RI, RD and RO for situation two where $\mathcal{N}=N(0,0.1)$. The tuple in the table is structured as (the score of MeTFA-smoothed explanation, the score of the original explanation)}
% \label{tab:normal insertion}
% \begin{tabular}{c c c c} 
% \toprule
% method & RI & RD & RO\\
% \midrule
% ScoreCAM & (1.292,0.943) & (0.428,0.404)& (0.864,0.539)\\
% RISE & (1.296,0.863) & (0.450,0.485)& (0.846,0.378) \\
% IGOS & (0.506,0.415) & (0.225,0.164)& (0.281,0.251) \\
% \bottomrule
% \end{tabular}
% \end{center}
% \end{table}

For the vanilla insertion, deletion and overall, the average scores of the $1000$ test images are shown in Table \ref{tab:insertion}. The bold digits in the table represent the higher faithfulness. It shows that, for ScoreCAM and RISE, two-sided MeTFA slightly decreases the value of insertion and increases the value of deletion, which suggests that two-sided MeTFA slightly reduces the faithfulness of the vanilla explanation algorithms. For IGOS, two-sided MeTFA slightly increases the value of insertion and increases the value of deletion. Thus, for such contradictory introduced in Section \ref{sec:setting}, overall is used to evaluate the faithfulness, and the results show that the two-sided MeTFA increases the faithfulness of IGOS. In general, the two-sided MeTFA maintains the faithfulness because the overall score is similar to the vanilla explanation.

For RI, RD and RO, the average scores of the $1000$ test images are shown in Table \ref{tab:normal insertion} where $\mathbb{O}_n=N(0,1)$. As we can see, the two-sided MeTFA significantly increases the robust faithfulness of the three vanilla explanations. Further experiments that calculate the RI, RD and RO with $\mathbb{O}_n=U(-0.1,0.1)$ confirm this conclusion and the results are shown in Table \ref{tab:uniform insertion} in the appendix. All of these results show that MeTFA significantly increases the robust faithfulness for the three explanations, regardless of $\mathbb{O}_n$ and $\mathbb{P}$ are the same or not.

As we can see, the vanilla RISE and ScoreCAM have higher overall score than MeTFA-smoothed ones. The reason could be that the explanations without MeTFA overfit the non-robust features or artifacts \cite{fong2017interpretable}, and thus receiving a higher faithfulness value, just as some models have higher accuracy on clean images but are more vulnerable to noises. MeTFA eliminates the effect of some non-robust features due to the sampling and the test. Thus, MeTFA slightly decreases the faithfulness of some explanation methods using the traditional metrics, but significantly increases the faithfulness using the proposed robust metrics.
\vspace{-1mm}
\subsubsection{Faithfulness in the Text Domain}
\begin{figure}
     \centering
     \includegraphics[width=\columnwidth]{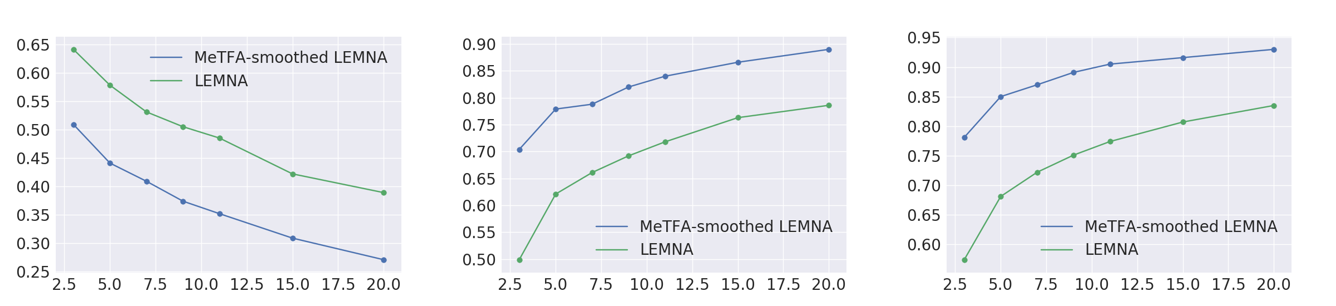}
     \caption{ The results of FDT, FAT and ST with different $n$ when $\mathbb{O}_n$ is $\mathbf{P}(0.5)$. The results of FDT, FAT and ST are shown from left to right, respectively. }
     \label{0_fidelity}
     
\end{figure}

\begin{figure}
     \centering
     \includegraphics[width=\columnwidth]{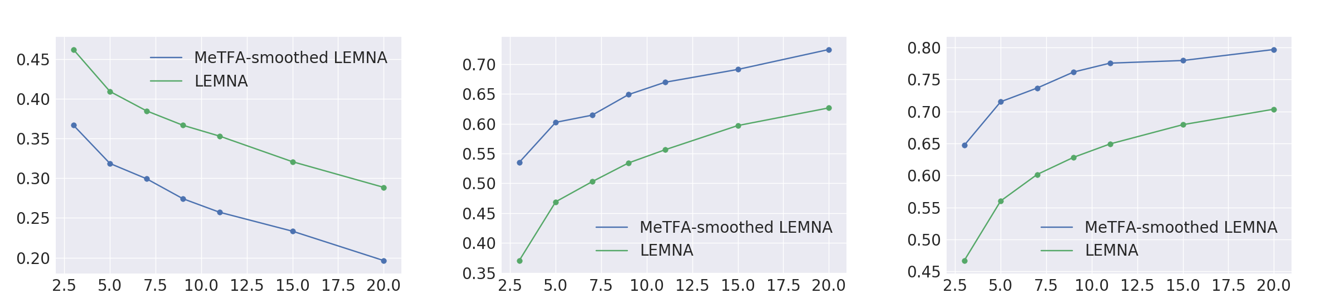}
     
     \caption{The results of of RFDT, RFAT and RST with different $n$ when $\mathbb{O}_n$ is $\mathbf{P}(0.5)$. The results of RFDT, RFAT and RST are shown from left to right, respectively.}
     \label{5_fidelity}
     
\end{figure}

In the text domain, we use FDT, FAT, ST and their corresponding robust metrics to estimate the faithfulness of an explanation. Similar to Section \ref{sec:stability}, we use synonym substitution to generate noise and set $\mathbb{P}$ and $\mathbb{O}_n$ to be $\mathbf{P}(0.5)$. As introduced in Section \ref{sec:setting}, the value of FDT changes with the number of processed features, \ie $n$. The results of FDT, FAT and ST with different $n$ are shown in Figure \ref{0_fidelity} for Toxic Comment dataset, where the number of samples of LEMNA is $2000$. It can be found that the FDT value (lower is better) of the MeTFA-smoothed LEMNA is always lower while the other two values (higher is better) are always higher, which suggests that the MeTFA-smoothed LEMNA is more faithful. The results of RFDT, RFAT and RST with different $n$ are shown in Figure \ref{5_fidelity} for Toxic Comment dataset, where the number of samples of LEMNA is $2000$. The results confirm that the two-sided MeTFA increases the faithfulness of LEMNA.
Further, we change the strength of the noise by setting $\mathbb{O}_n$ to $\mathbf{P}(0.3)$ and $\mathbf{P}(0.7)$. The results of the three robust faithfulness metrics are shown in Figure \ref{3_fidelity} and Figure \ref{7_fidelity} in the appendix, respectively, which consistently shows that MeTFA is better. Moreover, the results with another dataset (i.e., IMDb Reviews) and another number of samples for LEMNA (i.e., 500) are shown in Figure \ref{fig: total_fidelity_toxic_500} and Figure \ref{fig:total_fidelity_sentiment_2000} in the appendix, respectively. All of these results show that MeTFA can increase LEMNA’s faithfulness regardless of the existence of the real-world noises and the strength of the random noises.

% \ref{7_fidelity}. These results show that MeTFA can increase LEMNA's faithfulness regardless of whether there is random outer noise and the strength of the noise.

\vspace{-1mm}
\subsection{Impact of the Key Parameters}
In this part, we discuss the impact of several key parameters on the capabilities of MeTFA, including the threshold $h$ of the MeTFA-significant map, the number of sampled explanations $N$ and the confidence level $\alpha$. 
\vspace{-2mm}
\subsubsection{Threshold $h$}
Although we recommend to determine the threshold $h$ of the MeTFA-significant map by finding the optimal break, as discussed in Section \ref{sec:Significant}, $h$ is still a customizable parameter. In this part, we illustrate how $h$ influences the MeTFA-significant explanation and then show the advantage of applying the recommended threshold. 

Figure \ref{test h} shows the MeTFA-significant maps for RISE on an image of a macaque. The first row highlights the significantly important features, and the second row highlights the significantly unimportant features. When $h$ increases, the significantly important area becomes smaller, and the significantly unimportant area becomes larger, which is intuitive. To understand how well the highlighted area represents the model's prediction, we black out the image except the significantly important area and record the model's predicted score of the original label.  When $h=0.1$ and $0.2$, the significantly important region filters out the noisy features, leading to a higher predicted score compared to the original image. When $h= 0.3$ and $0.4$, the significantly important map keeps the predicted score with a smaller region. When $h=0.5$ and $0.6$, the area of the significantly important map is further reduced and the predicted score drops, but the prediction remains the same as the score is still larger than 0.5. Finally, when $h\ge 0.7$, the significantly important region is too small to keep enough information which causes a quick decrease of the predicted score. %\TBD{Add the case where h is calculated from 4.1}
% Thus, in the following application \ref{context}, we set $h=0.2$.\TBD{Dispute between me and MYH}
These phenomena show that the significantly important map correctly points out the features supporting the prediction of the model. In addition, when $h=0.8$, the significantly unimportant map covers almost the whole image but still gets a low  score, suggesting that the significantly unimportant map correctly points out the features which do not support the prediction of the model. 

By applying the optimal break method discussed in Section \ref{sec:Significant}, the recommended $h$ for this example is $0.4078$. Using this threshold, the significantly important map gets a high score with a small area. In addition, this threshold is almost the same to the score-area margin, $h=0.4$, as a smaller threshold keeps a much larger area and a higher threshold gets a much smaller predicted score. This example shows that the recommended method of determining the threshold is good for usage. Therefore, in the following applications (Section \ref{sec:context}), we use the recommended way to determine $h$ for the MeTFA-significant map. 
\vspace{-2mm}
\subsubsection{Number of Sampled Explanations $N$}
\begin{figure}[t]
    \centering
    \includegraphics[width=\columnwidth]{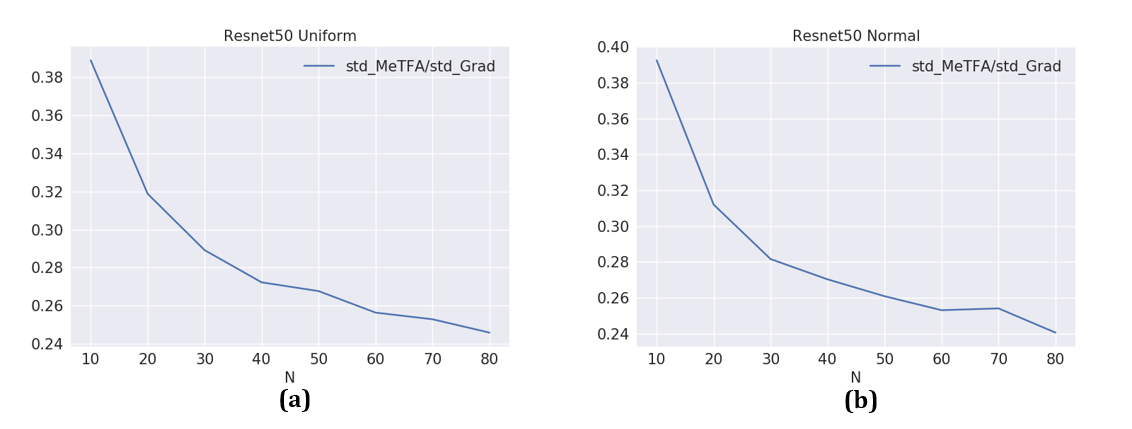}
    
    \caption{The results of $mstd_{MeTFA-Grad}/mstd_{Grad}$ for Resnet-50 when $N$ varies from $10$ to $80$.}
    \label{test repeat resnet}
    
\end{figure}

Although we give the lower bound for $N$ to achieve a confidence level $\alpha$ in Section \ref{sec: compute n}, $N$ is a customizable parameter as long as it is greater than the lower bound. In this part, we demonstrate how $N$ influences the stability of the MeTFA-smoothed explanation.
% As shown in \textbf{Theorem} \ref{th:MeTFA-able}, for one image, the variance of MeTFA-smoothed explanation has been proved to converge to zero with speed at least $O(1/N)$. 
As discussed in Section \ref{sec:stability}, we set the $\mathbb{P}$ and $\mathbb{O}_n$ to be the same and experiment with Normal and Uniform distributions. The results for the ResNet-50 are shown in Figure \ref{test repeat resnet}. As expected, the mstd of MeTFA-smoothed Gradient decreases when $N$ increases, meaning that the explanation is more stable with a larger $N$. Therefore, a user can custom $N$ according to the trade-off between the stability of the explanation and the tolerance of computational costs. However, even a small $N$, \eg, $N=10$, can bring significant stability benefits, as the std is reduced by over a half. %Further experiments testing on the Densenet169 suggest the similar conclusion, and the results are shown in the appendix (Figure \ref{test repeat densenet}).

\vspace{-2mm}
\subsubsection{Confidence Level $\alpha$}
$\alpha$ is a core parameter for hypothesis testing, and a lower $\alpha$ causes a more stringent test result. As discussed in Section \ref{sec: compute n}, a lower $\alpha$ requires a higher $N$. However, with a fixed large $N$, whether the choice of $\alpha$ makes a significant difference to the stability of the explanation remains unknown. To answer this practical question, we fix $N=50$ and experiment with different $\alpha$. Similar to the discussion of $N$, we test the stability of the MeTFA-smoothed Gradient for Resnet-50 by setting $\mathbb{P}$ and $\mathbb{O}_n$ both to be Normal and Uniform, respectively. The results are shown in Table \ref{Resnet alpha}. It shows that the stability does not change much with a smaller $\alpha$ when we reduce $\alpha$ from 0.05 to 0.0001. Therefore, MeTFA is insensitive to the value of $\alpha$. Further experiments on Densenet-169 imply the same conclusion, as shown in Table \ref{Densenet alpha} in the appendix. The intuition of this result is that $\alpha=0.05$ already implies the probability of noises to be tested significant is very small, and thus reducing it further does not benefit as much.
Therefore, we set $\alpha=0.05$ in the following application, which is common for a hypothesis test.

\begin{table}
\begin{center}
\caption{The results of $mstd_{MeTFA-Grad}/mstd_{Grad}$ for Resnet50 under two settings: $\mathbb{P}$ and $\mathbb{O}_n$ are both Normal or Uniform.}
\label{Resnet alpha}
\small
\begin{tabular}{c c c} 
\toprule
$\alpha$ & Uniform & Normal\\
\midrule
0.05 & 0.2558 & 0.2609\\
0.01 & 0.2553 & 0.2607\\
0.005 & 0.2552 & 0.2608\\
0.001 & 0.2552 & 0.2610\\
0.0005& 0.2554 & 0.2613\\
0.0001 & 0.2557 & 0.2619\\
\bottomrule
\end{tabular}
\end{center}

\end{table}

\vspace{-2mm}
\section{Application}
In this section, we apply  MeTFA to two applications closely related to security: detecting context bias in semantic segmentation and defending adversarial examples against the explanation-oriented attack. In this section, $\mathbb{P}$ is set to $U(-0.1,0.1)$.
% \subsection{Binary function start detection}
% Binary code reverse-engineering is to transfer binary code to assembly code. One of the key steps is to locate the start of binary function. Thus, we use the O1 dataset as the source dataset and the bi-directional RNN as the detecting model, both of which are provided by LEMNA \cite{guo2018lemna}. LEMNA can provide the most important $n$ features of the model for decision-making. Further, the authors defined three faithfulness test to validate the selected features. However, due to the randomness, the explanation of LEMNA is unstable, resulting in a low average faithfulness on the entire dataset, especially when the selected feature set is small. In this situation, we set the sample distribution of MeTFA only one data point, \ie $\mathbb{D}=\{x\}$ for every data $x$. \TBD{the table or image for quantitative comparison}

% TODO: continue here
\vspace{-1mm}
\subsection{Context Bias Detection}\label{sec:context}
As a component of the autonomous driving, semantic segmentation is of great importance. Formally, suppose that the class to be segmented is $c$, \eg, rider, and $I$ is the input image. A semantic segmentation model $F_c$ basically predicts for each pixel whether it belongs to $c$, represented by a probability score, and the segmentation result, denoted by $R$, is the set of all pixels that are predicted to be $c$.  An explanation  for the segmentation $R$ is an attribution map that highlights the area $R$ that the segmentation is based on. The highlighted area may include additional information that supports the prediction, \ie, $M\setminus R$. For example, it may highlight a bike that supports the segmentation of a rider, but is not included in the segmentation for the rider. This additional areas are called context bias for class $c$. Similar to the image classification task, an explanation for the segmentation is faithful if only keeping the highlighted areas is sufficient to produce the correct segmentation for $c$. Formally, we define the faithfulness value as follows:
\[ m_f = 1- \frac{Sum(Relu(R\odot (F_c(I)-F_c((R \cup M)\odot I))))}{Sum(R)} \] 
Basically, this metric measures how much the segmentation scores drop if we use only the explanation combined with the segmentation area to produce a new segmentation. A high $m_f$ means the model produces similar segmentation with the only area highlighted by the explanation. Similarly, we measure the robust faithfulness under noises sampled from $O_n$, defined by $E_{n\sim O_n} m_f(I+n)$.

\begin{figure}[t]
     \centering
     \includegraphics[width=\columnwidth]{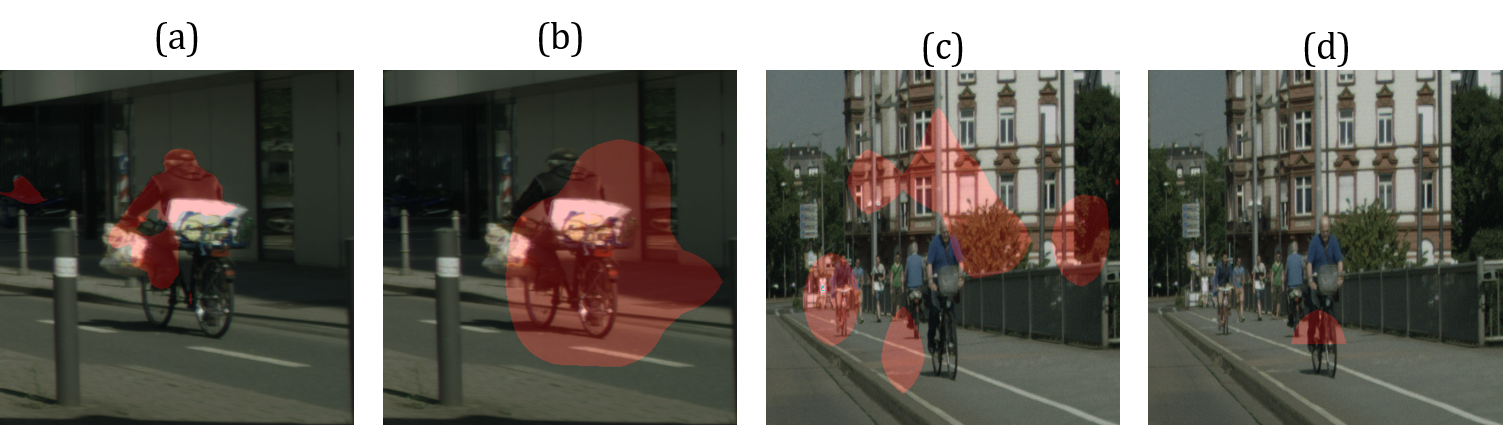}
     \caption{An example of the explanation for the semantic segmentation model. (a) and (b) are generated with the clean image. (a) is the \emph{rider} segmentation result and (b) is its explanation generated by GridSaliency. (c) and (d) are the explanation of the rider class with a noisy image. (c) is the original GridSaliency map, while (d) is the MeTFA-significant map.}
     \label{context_bias1}
     
\end{figure}

 GridSaliency \cite{hoyer2019grid} is the SOTA explanation algorithm for semantic segmentation. An example is shown in Figure \ref{context_bias1} (a) and (b). Figure \ref{context_bias1} (a) is the result of the model segmenting the rider class, and Figure \ref{context_bias1} (b) highlights the context bias of the rider class using GridSaliency. This result means that the model needs both the rider (Figure \ref{context_bias1} (a)) and the bike (Figure \ref{context_bias1} (b)) to recognize the rider. Using GridSaliency, engineers can debug a model when the model relies on the wrong context bias. However, the explanation of GridSaliency is vulnerable to random noise, which may mislead the practice. For example, after adding some noises to the original image, the attribution map for the rider becomes irrelevent, as shown in Figure \ref{context_bias1} (c).

% these two pictures tell that we need to put rider (Figure \ref{context_bias1}(a)) and bike (Figure \ref{context_bias1}(b)) into model so that model can identify rider, and in Figure \ref{context_bias2}(b), the attribution map tells that the model needs part of the building to segment rider. In fact, the context bias of rider and bicycle is consistent with intuition, while the rider and the building are not. Thus, users may think the former model is credible while the latter is not. However, both of the attribution maps are generated with the same model. The only difference between these two situation is that the former image is clean while the latter image is added with noise. The real world is full of noise. The explanation of GridSaliency is vulnerable to random noise, which may cause misjudgment by users. 
% the MeTFA-significant explanation can tell the significant important features which helps GridSaliency detect the context bias more accurately in the real world. 
% For example, the MeTFA-significant explanation tells that the bicycle is the significantly important region for the rider in this image. 

We apply the MeTFA-significant map to fix this issue. As shown in Figure \ref{context_bias1} (d), the MeTFA-smoothed map correctly highlights the bike again. We further  evaluate the faithfulness of the MeTFA-significant explanations using
the popular semantic segmentation dataset CityScapes \cite{Cordts2016Cityscapes} as the source dataset and the pre-trained PSPNet with R-50-D8 backbone \cite{mmseg2020} as the target model . We select three classes, tree, rider and car, as the target classes because intuitively rider has a strong context bias (bike) while trees and cars do not. For each class, we randomly select $100$ test images where the segmentation size is larger than $600$ pixels, to ensure that there exists at least one object segmented by the model as $c$ rather than some misclassified noisy pixels. To test the faithfulness of an explanation method in a noisy environment, we use noises sampled from $U(-0.1,0.1)$ to  compute the robust $m_f$.  For a fair comparison, we apply the same $h$ for the vanilla GridSaliency as well, \ie, we take the set of pixels with score higher than $h$ as the explanation of the vanilla GridSaliency. The results are shown in Table \ref{context_bias_table}. It can be seen that the MeTFA-significant explanations highlight far less pixels, e.g., about $2\%$ for trees, to maintain 99\%+ faithfulness, which suggests that the one-sided MeTFA filters out many noisy pixels in the vanilla GridSaliency map and keeps the pixels that the model really relies on. This can help engineers confidently determine if a model has context bias by looking at the the region the model significantly relies on.

Furthermore, from Table \ref{context_bias_table}, we can see that MeTFA filters out most of the context biases for the class tree and car, at 98\% and 92\%, respectively, while maintaining 40\% of the context bias for the class rider. This is intuitive because the model needs the context bias, \eg, bike, to determine whether a person is a rider. However, for trees and cars, the model does not need context bias to do so. Therefore, this results suggest that MeTFA is good at removing the false positives for the target classes and keeping only the correct context biases.
% For example, the area of highlighted pixels for the tree class in the MeTFA-significant map is only $1.98\%$ of that of the Gradsaliency map. As we illustrated before, the context bias of rider and bicycle (or motorcycle) is consistent with the intuition, and MeTFA-significant explanation points out that this context bias significantly exists. However, for the other classes, \ie, tree and car, the MeTFA-significant map highlights less than $10\%$ pixels to maintain $99+\%$ faithfulness. This phenomenon suggests that GridSaliency highlights some context bias while MeTFA-significant explanation shows that such context bias is not significant. 

In conclusion, the MeTFA-significant map can remove the noise of the attribution map and point out the context bias more accurately and confidently.
\begin{table}
\begin{center}
\caption{Evaluation of the highlighted area and the faithfulness of the attribution map for the segmentation model PSPNet. The second column is the faithfulness ratio of the MeTFA-significant map divided over the GridSaliency map. The third column is the ratio of the highlighted area.}
\label{context_bias_table}

\small
\begin{tabular}{c c c} 
\toprule
classes &$m_{MeTFA}/m_{GS}$&$\|M_{MeTFA}\|/\|M_{GS}\|$\\
\midrule
tree & 0.9983 & 0.0198  \\ 
rider & 0.9904 & 0.3955   \\
 car & 0.9915 & 0.080  \\
\bottomrule
\end{tabular}
\end{center}

\end{table}
% \begin{table}
% \begin{center}
% \begin{tabular}{||c | c| c||} 
% \hline
% classes &$m_{MeTFA}/m_{GS}$&$\|M_{MeTFA}\|/\|M_{GS}\|$\\
%  \hline\hline
%  tree & 0.9983984 & 0.0168436  \\ 
%  \hline
%  rider & 0.99105936 & 0.4102926   \\
%  \hline
%  car & 0.99135464 & 0.069932826  \\ [1ex] 
%  \hline
% \end{tabular}
% \end{center}
% \caption{Evaluation the area and the faithfulness of the attribution map. The second column is $m$ of MeTFA-significant map divided by $m$ of GridSaliency map. The third column is the area of MeTFA-significant map divided by the area of GridSaliency map.}
% \label{context_bias_table}
% \end{table}
\vspace{-1mm}
\subsection{Defending Explanation-Oriented Attacks}\label{attack CAM}

\begin{figure}[tbp]
\centering
\includegraphics[width=0.9\columnwidth]{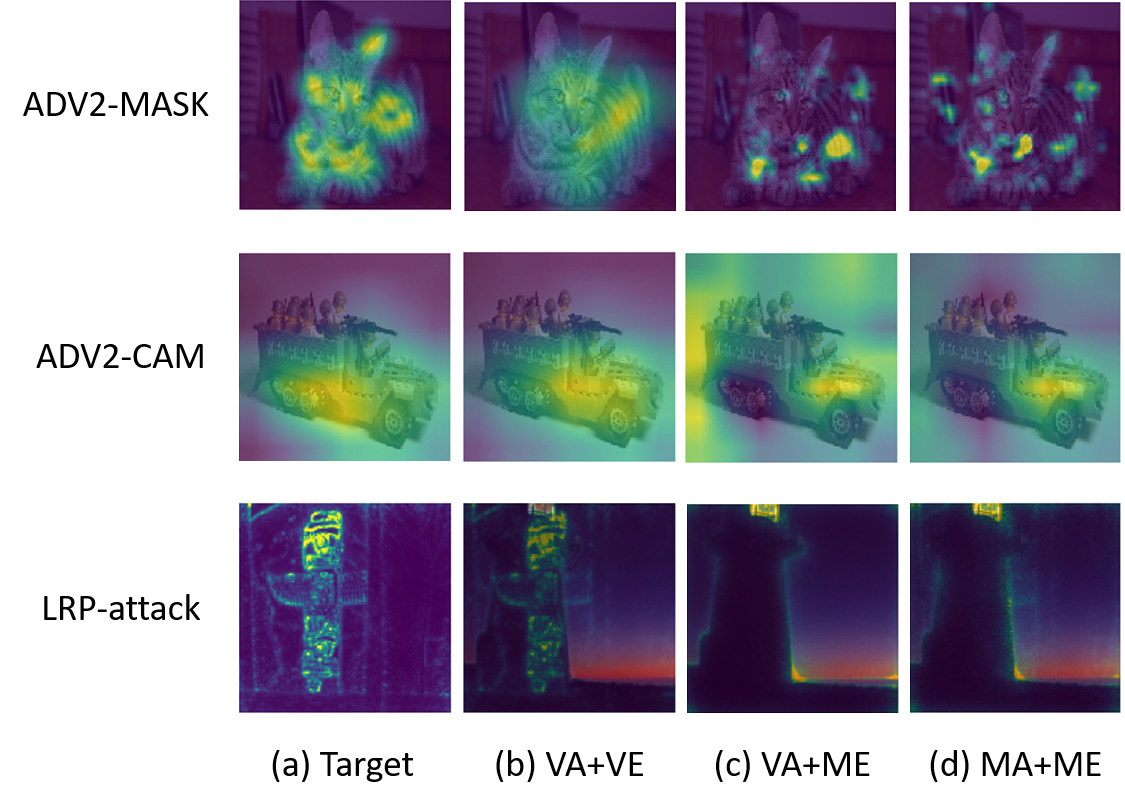} 
\caption{Examples of manipulation attack for MASK, CAM and LRP attack. (a) are the target maps. The attacker uses the vanilla attack (VA) to generate an adversarial example. Then the defender uses the vanilla explanation (VE) to generate a map for the adversarial example (b) or use the MeTFA-smoothed explanation (ME) to generate a map for the adversarial example (c). The attacker can also use adaptive attack against MeTFA (MA) to generate an adversarial example. Then the defender uses the MeTFA-smoothed explanation (ME) to generate a map for the adversarial example (d).}
\label{fig:ada_attack} 
\end{figure}

Explanations are designed to help human understand and trust the model. However, recent works show that the explanation can be manipulated. Manipulation attack \cite{dombrowski2019explanations} is able to keep the model's prediction unchanged, but manipulate the attribution maps generated by the explanation algorithms arbitrarily. As shown in the second row of Figure \ref{fig:ada_attack} (a) and (b), an attacker can manipulate the explanation to be similar to a target map. Moreover, some AI systems use feature attribution to detect adversarial explanations, but ADV$^2$ \cite{zhang2020interpretable} can evade such detection by manipulating the adversarial example's explanation to be similar to the benign one. As shown in the first row of Figure \ref{fig:ada_attack} (a) and (b), an attacker changes the predicted label of the image from check to sandbar while manipulating its explanation (b) to be similar to the benign one (a). ADV$^2$ attack can be decomposed into two steps: an attacker first attacks the prediction of the model only and then manipulates the explanation to the benign one while maintaining the target label. Therefore, the core of the above two attacks is the same, i.e., manipulate the explanation to a target map while maintaining the predicted label. 

We conduct experiments to test MeTFA's ability to defend the explanation-oriented attack quantitatively. Similar to Section \ref{sec:dataset and model}, we use ILSVRC2012 val as the source dataset. We test the attack on MeTFA-smoothed CAM \cite{zhang2020interpretable} (a CAM -based explanation), MeTFA-smoothed LRP\cite{dombrowski2019explanations} (a gradient-based explanation) and MeTFA-smoothed MASK \cite{zhang2020interpretable} (an optimization-based explanation). Since the existing attack methods rely on the gradient relation between the attribution map and the input image, they could not attack the sample-based explanations, where no gradient information is available. All the attack methods  follow the default settings of the original paper. 

The aim of the attacker is to manipulate the explanation to a target pattern while keeping the predicted label. Correspondingly, the aim of the defender is to make the generated explanation different from the target pattern. We evaluate the defense capabilities of MeTFA from two aspects: visual effects and quantitative analysis. For each aspect, we conduct experiments for both conditions where the attacker knows or does not know MeTFA. Formally, the objective functions to optimize the adversarial example for the original attack (Equation \ref{pc:original attack}) and the adaptive attack (Equation \ref{pc:adaptive attack}) are as follows:
\vspace{-2mm}
\begin{equation}\label{pc:original attack}
    x = \arg \min_{x}({\lambda_1\|f_c(x)-f_c(x_0)\|+\lambda_2\|x-x_0\|+\lambda_3{\|E(f_c,x)-E_t\|}})
\end{equation}
\begin{equation}\label{pc:adaptive attack}
\begin{aligned}
    x =& \arg \min_{x}(\lambda_1\|f_c(x)-f_c(x_0)\|+\lambda_2\|x-x_0\| +\lambda_3{\|E(f_c,x)-E_t\|}\\&+\lambda_4\|E_M(f_c,x)-E_t\|)
\end{aligned}
\end{equation}
where $f_c(x)$ returns the predicted score of class $c$, $x_0$ is the original data, $E(f_c,x)$ returns the vanilla attribution map of $x$, $E_t$ is the target attribution map and $E_M(f_c,x)$ returns the MeTFA-Smoothed attribution map of $x$.
In the experiment, we consider the strongest adaptive attacker, i.e., the attacker uses the hyperparameters exactly the same as the defender who uses MeTFA to denfend against the adversarial examples.

First, the visual results of MeTFA are shown in Figure \ref{fig:ada_attack} (c) and (d) for MeTFA-smoothed explanation of the vanilla adversarial examples and that of the adaptive adversarial examples. As we can see, whether the attacker knows MeTFA, the MeTFA-smoothed explanations are very different to the targets, which suggests the ability of MeTFA to defend against adversarial examples in practice.

Second, we quantify the ability of the MeTFA to defend against the attack. Formally, we denote the target map as $m_1$ (e.g., Figure \ref{fig:ada_attack}(a)), the map generated with vanilla explanation for the vanilla adversarial example as $m_2$ (e.g., Figure \ref{fig:ada_attack}(b)), the map generated with MeTFA-smoothed explanation for the vanilla adversarial example as $m_3$ (e.g., Figure \ref{fig:ada_attack}(c)) and the map generated with MeTFA-smoothed explanation for the adaptive adversarial example as $m_4$ (e.g., Figure \ref{fig:ada_attack}(d)). We define the distance between two maps as 
$ d(m,n) = \frac{1}{|S|}\sum_{(i,j)\in S}|m_{ij}-n_{ij}| $,
where $m_{ij}$ is the value of the pixel $(i,j)$ in $m$. To quantify the difference between these maps, we test  $100$ random selected images from ImageNet and show the average distance of the $100$ images in Table \ref{tab:ada_attack}. When the attacker is not adaptive, we can see that $d(m_1,m_3)$ is significantly greater than $d(m_1,m_2)$, which suggest MeTFA can defend against such attack. When the attacker is adaptive for MeTFA, we can see $d(m_1,m_4)$ is still significantly greater than $d(m_1,m_2)$. Therefore, MeTFA can weaken the attacker’s ability to manipulate the explanation even when the attacker knows MeTFA.

\begin{table}[tbp]
\begin{center}
\caption{The distance between the maps for CAM, LRP and MASK.}
\label{tab:ada_attack}
% \vspace{0.2cm}
\begin{tabular}{cccc}
\toprule
                    %   &                       & \multicolumn{4}{c}{LIME}               \\
 & $d(m_1,m_2)$ & $d(m_1,m_3)$ & $d(m_1,m_4)$    \\
\midrule
 CAM            & 0.0934 & 0.2785  & 0.1705\\
 LRP            & 0.0340 & 0.0657  & 0.0635\\
 MASK           & 0.1453 & 0.1667  & 0.1648\\
\bottomrule
\end{tabular}
\end{center}
\end{table}

\vspace{-2mm}
\section{Discussion}\label{sec:discuss}

\paragraphbe{Bonferoni correction}
Bonferoni correction guides researchers to use union bounds when computing confidence interval. We do not do a Bonferoni correction for computing confidence intervals over multiple features. Instead, MeTFA define a two-fold hypothesis testing to generate confidence interval for each feature individually. This is because Bonferoni correction requires a lot more queries for high dimensional data (e.g., a 224$\times$224-dimensional image) to get tighter bounds and thus is inefficient in practice. For efficiency, we consider the confidence interval for each feature individually and the experiments show it works well.

% \begin{figure}[t]
% \centering
% \includegraphics[width=0.9\columnwidth]{figures/birdhouse.png}
% \caption{The IGOS explanation of an image whose predicted class is birdhouse. (a) and (b) are the visualizations of the explanations. (c) is the image after inserting $50100$ pixels.}
% \label{fig:birdhouse}
% \end{figure}

\paragraphbe{Asymptotic Quantification from Extending SmoothGrad}

\noindent Although SmoothGrad does not quantify the uncertainty of explanation, it can be extended to provide an asymptotic confidence bound. Using Jackknife method \cite{wiki:Jackknife_resampling}, one can estimate the standard deviation $\sigma$ of the smoothed explanation $\mu$. By the central limit theorem \cite{wiki:Central_limit_theorem}, the smoothed explanation asymptotically converges to a normal distribution $\mathcal{N}(\mu, \sigma^2)$. Therefore, the $\alpha$ confidence bound is $[\mu-p_{\alpha/2}\sigma, \mu+p_{\alpha/2}\sigma]$, where $p_{\alpha/2}$ is the upper $\alpha/2$ quantitle of the standard normal distribution. However, this bound is only valid when $N$ is large, while our bound is valid for all $N$.

\paragraphbe{Limitation of MeTFA} 
As illustrated in Section \ref{sec:faithful}, if an explanation algorithm produces unfaithful explanations due to the effectiveness of non-robust features or the randomness in the algorithm, then MeTFA can significantly help and produce better explanation. However, if the explanation algorithm has major problems, such as attributing all features with the same value, then MeTFA can hardly generate faithful explanations. In other words, MeTFA can only make weak explanations stronger, by removing noises in the attribution maps.
\vspace{-2mm}
\section{Conclusion}
In this paper, we propose MeTFA, the first work to quantify and reduce the randomness in feature attribution methods with theoretical guarantees. 
% By conducting one-sided test and two-sided test, MeTFA can estimate and reduce the randomness in any feature attribution method.
By evaluating with extensive experiments, we show that MeTFA can increase the stability of explanation while maintaining the faithfulness. With the proposed robust faithfulness metrics, we show that MeTFA-smoothed explanations significantly increase the explanation's ability to locate robust features. In addition, we demonstrate that MeTFA can detect context bias in the semantic segmentation model more accurately and defend against the explanation-oriented attack, which shows its great potential in practice.
\vspace{-2mm}
\section{ACKNOWLEDGMENTS}
We would like to gratefully thank the anonymous reviewers for their helpful feedback. This work was partly supported by the Zhejiang Provincial Natural Science Foundation for Distinguished Young Scholars under No. LR19F020003, NSFC under No. 62102360, U1936215, and U1836202, and the Open Research Projects of Zhejiang Lab under No. 2022RC0AB01. Ting Wang is partially supported by the National Science Foundation under No. 1951729, 1953893, 2119331, and 2212323.

\bibliographystyle{ACM-Reference-Format}
% \bibliography{ccs-sample}
\bibliography{reference.bib}

\clearpage
\appendix
\section{Appendix}

\subsection{Proofs}
\label{sec:proof}

\subsubsection{Proof of Proposition \ref{prop:one-sided-test}}\label{prof:prop:one-sided-test}

We only prove the case when $h\ge V$. The other case can be derived similarly. Since $ct_j(h) := \sum_{i=1}^N I(e_{ij}\ge h)$ is the sum of i.i.d. Bernoulli variables, for a fixed $q_j(h):=P(e_{ij}\ge h)$ we can easily obtain $P(ct_j(h)\ge m)$. The problem here is that we do not know $q_j(h)$, except that $q_j(h)\ge 0.5$ for $h\ge V$. Therefore, we can get the upper bound of $P(ct_j(h)\ge m)$ via $\max_{q_j(h)\ge 0.5} P(ct_j(h)\ge m)$.

We have known that $ct_j(h)\sim B(p)$, where $p=q_j(h)\le q_j(V)=0.5$. Therefore, $P(ct_j(h)=m)= \binom{N}{m}\times p^m \times (1-p)^{N-m}$. For a fixed $m$, the probability increases monotonically for $p\in [0,m/N)$ and decreases monotonically for $p\in (m/N,1]$. Thus the probability has and only has one maximal, achieved by $p_*=\min(0.5, m/N)$. Therefore, $P(ct_j(h)=m)\le \binom{N}{m}\times p_*^m \times (1-p_*)^{N-m}$ for every $m$. Adding up this inequality w.r.t. $m$ proves the statement.

\subsubsection{Proof of Theorem \ref{th:one-sided-MeTFA}}\label{prof:th:one-sided-MeTFA}
For $H_0:V_j\le h$, the $p$-value is the probability of abnormal event $P(ct_j(h)\ge m)$ with a large $m$. Therefore, according to Proposition \ref{prop:one-sided-test}, its $p$-value is less than or equal to $\sum_{i=k}^N \binom{N}{i} \times p_*^i \times (1-p_*)^{N-i}$. Similarly, for $H_0:V_j\ge h$, the $p$-value is the probability of abnormal event $P(ct_j(h)\le m)$ with a small $m$. Therefore, according to Proposition \ref{prop:one-sided-test}, its $p$-value is less than or equal to $\sum_{i=0}^k \binom{N}{i} \times p_*^i \times (1-p_*)^{N-i}$.

\subsubsection{Proof of Proposition \ref{prop:two-sided-test}}\label{prof:prop:two-sided-test}
Since $q_j(h)=0.5$ and $ct_j(h) \sim B(q_j(h))$, we have $P(ct_j(h)\le m_1) \le \sum_{i=0}^{m_1} \binom{N}{i} \times 0.5^N$ and $P(ct_j(h)\ge m_2) \le \sum_{i=m_2}^N \binom{N}{i} \times 0.5^N$. Adding them together completes the proof.

\subsubsection{Proof of Theorem \ref{th:two-sided-MeTFA}}\label{prof:th:two-sided-MeTFA}
For $H_0:V_j= h$, the $p$-value is the probability of abnormal event $P(ct_j(h)\ge m)$ with a large $m$ and $P(ct_j(h)\le m)$ with a small $m$. Therefore, according to Proposition \ref{prop:two-sided-test}, its $p$-value is less than or equal to $\sum_{i\in \{0,\dots,k_1\} \cup \{k_2,\dots,N\}} 0.5^N \times \binom{N}{i}$.

\subsubsection{Proof of Theorem \ref{th:interval}}\label{prof:th:interval}
By Proposition \ref{prop:two-sided-test}, under $H_0: V_j=h_0$, $P_{h_0}(ct_j(h_0)\in (k_{1}, k_{2})) = 1-P(ct_j(h_0)\le k_1) - P(ct_j(h_0)\ge k_2) = 1-2 \times P(ct_j(h_0)\le k_1) \ge 1-2\times \sum_{i=0}^{k_1} 0.5^N \times \binom{N}{i} \ge 1-\alpha$. By the arbitrariness of $h_0$, we can write $P_{h}(ct_j(h)\in (k_{1}, k_{2}))\ge 1-\alpha$ for any $h$. Furthermore, the defined $h_{1j}$ and $h_{2j}$ satisfies: $h_{1j}=\argmin_h \{ct_j(h) \in (k_{1}, k_{2})\}$ and $h_{2j}=\argmax_h \{ct_j(h) \in (k_{1}, k_{2})\}$. Thus, $ct_j(h)\in (k_{1}, k_{2})$ is equivalent to $h \in [h_{1j}, h_{2j}]$ and $P_{h}(h \in [h_{1j}, h_{2j}])\ge 1-\alpha$. Therefore,  $[h_{1j}, h_{2j}]$ is a $1-\alpha$ confidence interval for $V_j$.

\subsubsection{Proof of Theorem \ref{th:MeTFA-able}}\label{prof:th:MeTFA-able}

(1) First, we prove the convergence of $\ST_j$. The main idea is to use the normal approximation of the Bernoulli distribution which is exact when $N \rightarrow \infty$, so that we can explicitly write $k_1$ and $k_2$ in terms of $N$ and $\alpha$. Then we take the limit $N \rightarrow \infty$ to conclude both $k_1$ and $k_2$ converge to $N/2$. Therefore, it follows that the average of samples between the $k_1$ and $k_2$ index converges to the median.

By applying the normal approximation of Bernoulli distribution, we have $\alpha/2 \approx \sum_{i=0}^{k_1} 0.5^N \binom{N}{i} \approx F_{\N(N/2,N/4)}(k_1)$ for large $N$, where $F_{\N(N/2,N/4)}$ is the CDF of normal distribution with mean $N/2$ and variance $N/4$. In addition, we can write $F_{\N(N/2,N/4)}(k_1)=F_{\N(0,N/4)}(k_1-\frac{N}{2})=F_{\N(0,1)}((k_1-\frac{N}{2})/(\frac{\sqrt{N}}{2}))$. Therefore, $(k_1-\frac{N}{2})/(\frac{\sqrt{N}}{2}) = -\mu_{\alpha/2}$, where $\mu_{\alpha/2}$ is the $\alpha/2$ upper quantile of normal distribution. By rearranging this equation, we have $k_1\approx -\frac{\sqrt{N}}{2}\mu_{\alpha/2}+\frac{N}{2}$. Thus, $k_2=N-k_1\approx \frac{N}{2}+\frac{\sqrt{N}}{2}\mu_{\alpha/2}$. Therefore, $k_1/N$ and $k_2/N$ converges to $1/2$. Now, $\ST_j=\sum_{i=k_1+1}^{k_2-1} e_{(i)j}/(k_2-k_1-1) \ge \sum_{i=k_1+1}^{k_2-1} e_{(k_1)j}/(k_2-k_1-1) = e_{(k_1)j} \rightarrow e_{(N/2)j} = V_j$ when $N\rightarrow \infty$. Similarly, $\ST_j=\sum_{i=k_1+1}^{k_2-1} e_{(i)j}/(k_2-k_1-1) \le \sum_{i=k_1+1}^{k_2-1} e_{(k_2)j}/(k_2-k_1-1) \rightarrow V_j$ when $N\rightarrow \infty$. Using the pinching theorem, we get $\ST_j\rightarrow V_j$ when $N\rightarrow \infty$.

(2) Second, we compute the lower bound of the convergence rate for $\ST_j$. The main idea is to upper bound the variance of $\ST_j$, and then show the upper bound decreases in the speed of $O(1/N)$.

We will apply the \emph{approximate summaries} theorem from Page 120 of Baglivo \cite{baglivo_2005}, which states that for any $X_i$ sampled from some distribution, $\Var(e_{(k)j})\approx \frac{p(1-p)}{(N+2)(f_e(\theta)^2)}$ where $p=\frac{k}{N+1}$, $\theta$ is the $p$th lower quantile and $f_e$ is the PDF of $e$. Applying this theorem directly to compute the variance of $e_{(k_1+i)j}$ for $i$ from $1$ to $k_2-k_1-1$, we get $\Var(e_{(k_1+i)j})\approx \frac{(k_1+i)(N-k_1+1-i)}{(N+1)^2(N+2)}\times \frac{1}{f_e{(\theta)}}\approx \frac{(\frac{N}{2}-\frac{\sqrt{N}}{2}\mu_{\alpha/2}+i)(\frac{N}{2}+\frac{\sqrt{N}}{2}\mu_{\alpha/2}-i)}{(N+1)^2(N+2)}\times \frac{1}{f_e{(\theta)}}$. Since $i$ is in the order of $O(\sqrt{N})$, we have $\Var(e_{(k_1+i)j})\approx \frac{1}{4N} \times \frac{1}{f_e{(\theta)}}$. In addition, $p=\frac{k_1+i}{N+1}\rightarrow \frac{1}{2}$, which means $\theta=V_j$. Therefore, under the mild assumption that $f_e(V_j)>0$, $\frac{1}{f_e{(\theta)}}$ converges to a constant. Combining this fact with the previous formula, we have $Var(e_{(k_1+i)j})=O(1/N)$. Since $\Var(A+B)=\Var(A)+\Var(B)+2\Cor(A,B)\le (\sqrt{\Var(A)}+\sqrt{\Var(A)})^2$ for any $A$ and $B$, using induction, we get $\Var(\sum_{i=1}^{k} X_i) \le (\sum_{i=1}^{k} \sqrt{\Var(X_i)})^2$. By applying this formula on $\ST_j$, we get
$
	\Var(\ST_j) \approx \frac{1}{\mu_{\alpha/2}^2 N} \Var( \sum_{i=k_1+1}^{k_2-1} e_{(i)j}) \le  \left(O(\frac{1}{\sqrt{N}})\times O(\sqrt{N})\right)^2 \times O(\frac{1}{N}) = O(\frac{1}{N}).
$

% \begin{align}
% 	\Var(\ST_j) &\approx \frac{1}{\mu_{\alpha/2}^2 N} \Var( \sum_{i=k_1+1}^{k_2-1} e_{(i)j})\\& \le O(1/N) \times \left(O(1/\sqrt{N})\times O(\sqrt{N})\right)^2 = O(1/N)
% \end{align}

\subsection{Additional Experimental Results}

\subsubsection{Examples}

(1) An example to show the advantage of MeTFA compared to LEMNA.
We provide an example predicted by the model as toxic as follows. The red words are the top 3 important words found by the attribution map, and the yellow words are the top 4-7 important words. As we can see, this text is misclassified by the model, and MeTFA tells that `Dick' misleads the model while the vanilla LEMNA fails to find this. 

LEMNA: "\textcolor{red}{I'm} \textcolor{yellow}{pretty} sure this is a joke. The other books for sale are How to raise a Jewish dog and Yiddish for Dick \textcolor{yellow}{and} \textcolor{yellow}{Jane}. It is an expensive and \textcolor{yellow}{elaborate} hoax. And, after the novelty wears off, not even \textcolor{red}{that} \textcolor{red}{funny}."

MeTFA-smoothed LEMNA:"I'm pretty sure this is a joke. The other books for sale are How to raise a Jewish dog and Yiddish for \textcolor{red}{Dick} \textcolor{red}{and} \textcolor{yellow}{Jane}. \textcolor{yellow}{It} is an expensive and \textcolor{yellow}{elaborate} hoax. And, after the novelty wears off, not even \textcolor{yellow}{that} \textcolor{red}{funny}."

\begin{figure*}[htbp]
    \centering
    \includegraphics[width=0.9\textwidth]{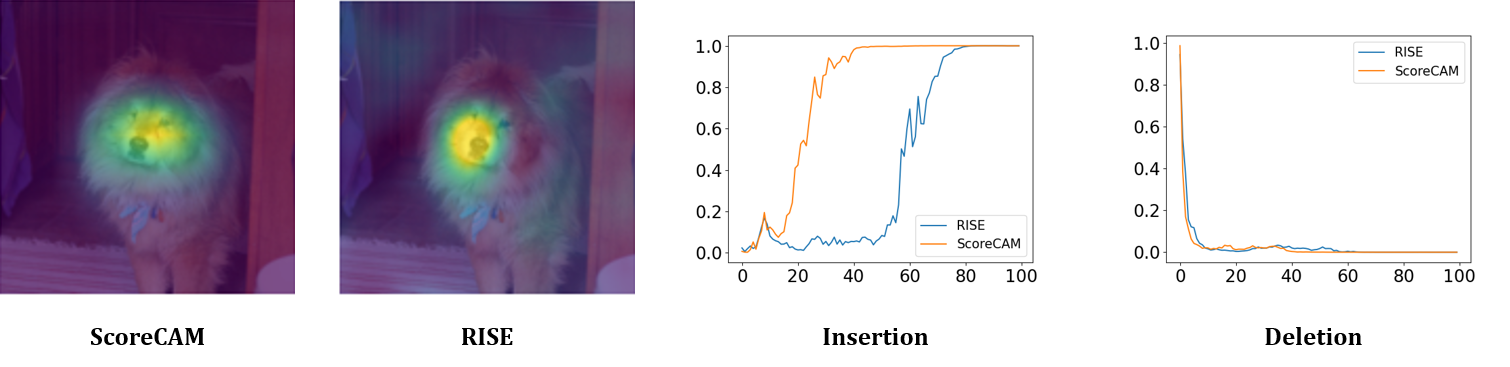}
    \caption{An example of the insertion metric and the deletion metric. }
    \label{fig:insertion and deletion}
\end{figure*}

(2) An example to show how the insertion and deletion calculate.
The example is shown in \ref{fig:insertion and deletion}. For insertion, the chow chow's predicted score increases while inserting the top $n$\% important pixels highlighted by the attribution map, where $n$ from $0$ to $100$. A more faithful attribution map highlights the supportive features more accurately, thus getting a higher area under the curve. On the opposite, we gradually delete the important pixels, and hence the predicted score of chow chow is reducing. A more faithful attribution map should have a lower area under the deletion curve. Thus, we can use the area under the insertion curve or deletion curve to reflect the accuracy of the attribution map to highlight the supportive features.

% \subsubsection{Tables}
\begin{table}[htp]
\begin{center}
\caption{The mstd value of the MeTFA-smoothed Gradient and the vanilla Gradient for Resnet50.}
\label{tab:stability Gradient Resnet}
\begin{tabular}{c c c c c} 
\toprule
$\mathbb{P}\backslash\mathbb{N}$& Normal & Uniform &Brightness& Avg\\
\midrule
Normal & \textbf{0.0176} & \textbf{0.0154} & 0.0142 & \textbf{0.0157}\\
Uniform & 0.0208 & 0.0171 & 0.0142 & 0.0173\\
Brightness & 0.0409 & 0.0403 & \textbf{0.0104} & 0.0305\\
\midrule
Vanilla Gradient & 0.0457 & 0.0445 & 0.0170 & 0.0357\\
\bottomrule
\end{tabular}
\end{center}
\end{table}

\begin{table}[htp]
\begin{center}
\caption{The mstd value of the MeTFA-smoothed LIME and the vanilla LIME for Resnet50. }
\label{tab:LIME_std}
\begin{tabular}{ccccc}
\toprule
 $\mathbb{P}$\textbackslash{}$\mathbb{O}_n$ & Normal & Uniform & Brightness & Avg    \\
\midrule
 Normal                & \textbf{0.0868} & \textbf{0.0752}  & 0.074      & \textbf{0.0786} \\
 Uniform               & 0.0907 & 0.0761  & 0.072      & 0.0796 \\
 Brightness            & 0.1524 & 0.1486  & \textbf{0.0698}     & 0.1236 \\
\midrule
Vanilla             & 0.2148 & 0.1979  & 0.1193     & 0.1773\\
\bottomrule
\end{tabular}
\end{center}
\end{table}

\begin{table}[htp]
\begin{center}

\caption{The mstd value of the MeTFA-smoothed IGOS and the vanilla IGOS for Resnet50. }
\label{tab:IGOS_std}
\begin{tabular}{ccccc}
\toprule
                    %   &                       & \multicolumn{4}{c}{IGOS}                \\
$\mathbb{P}$\textbackslash{}$\mathbb{O}_n$ & Normal & Uniform & Brightness & Avg     \\
\midrule
 Normal                & \textbf{0.0278} & \textbf{0.0202}  & 0.0156     & \textbf{0.0212}  \\
 Uniform               & 0.0308 & 0.0214  & 0.0131     & 0.02176 \\
 Brightness            & 0.0341 & 0.0258  & \textbf{0.0093}     & 0.0231  \\
\midrule
Vanilla             & 0.0437 & 0.0335  & 0.0184     & 0.0318 \\
\bottomrule
\end{tabular}
\end{center}
\end{table}

\begin{table}[htp]
\begin{center}

\caption{The mstd value of the MeTFA-smoothed ScoreCAM and the vanilla ScoreCAM for Resnet50. }
\label{tab:ScoreCAM_std}
\begin{tabular}{ccccc}
\toprule
                    %   &                       & \multicolumn{4}{c}{ScoreCAM}           \\
 $\mathbb{P}$\textbackslash{}$\mathbb{O}_n$ & Normal & Uniform & Brightness & Avg    \\
\midrule                       
 Normal                & \textbf{0.0469} & \textbf{0.033}   & 0.0201     & \textbf{0.033}  \\
 Uniform               & 0.0522 & 0.0344  & 0.0168     & 0.0341 \\
 Brightness            & 0.0583 & 0.0458  & 0.012      & 0.0387 \\
\midrule
Vanilla                   & 0.0607 & 0.0465  & \textbf{0.0112}     & 0.0395\\
\bottomrule
\end{tabular}
\end{center}
\end{table}

\begin{table}[htp]
\begin{center}
\caption{The mstd value of the MeTFA-smoothed Gradient and the vanilla Gradient for Densenet169. }
\label{tab:stability Gradient Densenet}
\begin{tabular}{c c c c c} 
\toprule
$\mathbb{P}\backslash\mathbb{O}_n$& Normal & Uniform &Brightness &Avg\\
\midrule
Normal & \textbf{0.0184} & \textbf{0.0164} & 0.0155 & \textbf{0.0167} \\
Uniform & 0.0217 & 0.0185 & 0.0155 & 0.0186 \\
Brightness & 0.0457 & 0.0445 & \textbf{0.0095} & 0.0332 \\
\midrule
Vanilla Gradient & 0.0495 & 0.0489 & 0.0140 & 0.0374 \\
\bottomrule
\end{tabular}
\end{center}
\end{table}

\begin{table}[htp]
\begin{center}
\caption{The results with the robust faithfulness metric with $\mathbb{O}_n=N(0,0.1)$. The tuple in the table is structured as (the score of MeTFA-smoothed explanation, the score of the vanilla explanation).}
\label{tab:uniform insertion}
\begin{tabular}{c c c c} 
\toprule
method & RI & RD & RO\\
\midrule
ScoreCAM & (\textbf{0.7718},0.6187) & (\textbf{0.2024},0.2068)& (\textbf{0.5694},0.4119)\\
RISE & (\textbf{0.6600},0.4738) & (0.2645,\textbf{0.1620})& (\textbf{0.3955},0.3118) \\
IGOS & (\textbf{0.4295},0.3288) & (0.0953,\textbf{0.0932})& (\textbf{0.3342},0.2356) \\
\bottomrule
\end{tabular}
\end{center}
\end{table}

% %Densenet169 outer=inner= uniform or normal
\begin{table}[htp]
\begin{center}
\caption{The results of $\emph{std}_{MeTFA-Grad}/\emph{std}_{SG-Grad}$ for Densenet169 when $\mathbb{P}$ and $\mathbb{O}_n$ are both Uniform or Normal.}
\label{Densenet vs_sg divide}
\begin{tabular}{c c c} 
\toprule
Sampling number & Uniform & Normal\\
\midrule
10 & 0.9550 & 0.9473\\
30 & 0.9280 & 0.9175\\
50 & 0.9171 & 0.9052\\
70 & 0.9093 & 0.8985\\
\bottomrule
\end{tabular}
\end{center}
\end{table}

\begin{table}[htp]
\begin{center}
\caption{The results of $mstd_{MeTFA-Grad}/mstd_{SG-Grad}$ for VGG16 when $\mathbb{P}$ is Uniform and Normal.}
\label{vgg vs_sg divide outer_none}
\begin{tabular}{c c c} 
\toprule
$N$ & Uniform & Normal\\
\midrule
10 & 0.9361 & 0.9237\\
30 & 0.9374 & 0.9141\\
50 & 0.9478 & 0.9241\\
70 & 0.9603 & 0.9345\\
\bottomrule
\end{tabular}
\end{center}
\end{table}

\begin{table}[htp]
\begin{center}
\caption{The results of $\emph{std}_{MeTFA-Grad}/\emph{std}_{Grad}$ for Densenet169 under two settings: $\mathbb{P}$ and $\mathbb{O}_n$ are both Uniform or Normal.}
\label{Densenet alpha}
\begin{tabular}{c c c} 
\toprule
$\alpha$ & Uniform & Normal\\
\midrule
0.05 & 0.2558 & 0.2441\\
0.01 & 0.2553 & 0.2436\\
0.005 & 0.2552 & 0.2436\\
0.001 & 0.2552 & 0.2437\\
0.0005& 0.2554 & 0.2439\\
0.0001 & 0.2557 & 0.2442\\
\bottomrule
\end{tabular}
\end{center}
\end{table}

\begin{table}[htp]
\begin{center}
\caption{The results for Toxic Comment dataset. The mstd values for LEMNA(500), MeTFA-smoothed LEMNA(500), smoothed LENMNA(500) when $\mathbb{P}$ and $\mathbb{O}_n$ are both $\mathbf{P}(0.5)$.}
\label{tab:lemna vs metfa vs sg}
% \vspace{0.2cm}
\small
\resizebox{0.9\linewidth}{!}{
\begin{tabular}{c  c c c} 
\toprule
noise & mstd of LEMNA & mstd of MeTFA-smoothed LEMNA & mstd of smoothed LEMNA\\
\midrule
$\mathbf{P}(0.5)$ & 0.1531 & \textbf{0.05488} & 0.0667\\
\bottomrule
\end{tabular}
}
\end{center}
% \vspace{-0.1cm}
\end{table}

% \clearpage
% \subsubsection{Figures}
\begin{figure}[htp]
     \centering
     \includegraphics[width=\columnwidth]{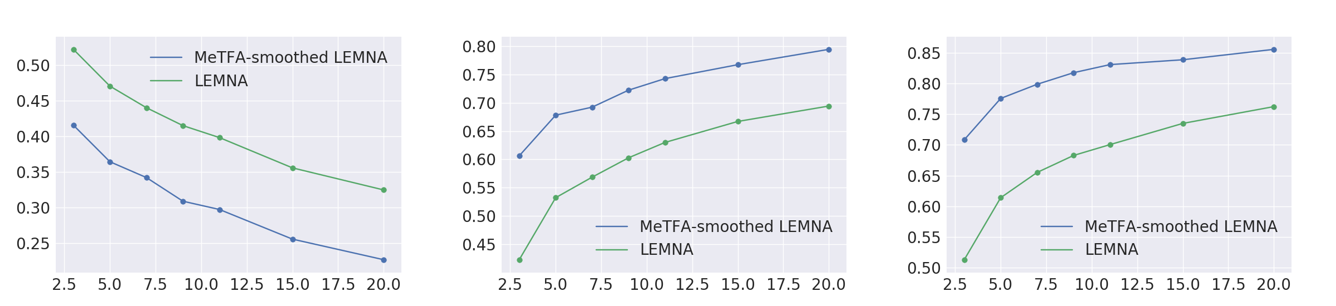}
     \caption{The results of of RFDT, RFAT and RST with different $n$ where $\mathbb{O}_n$ is $\mathbf{P}(0.3)$. From left to right, there are the results of RFDT, RFAT and RST, respectively. }
     \label{3_fidelity}
\end{figure}

\begin{figure}[htp]
     \centering
     \includegraphics[width=\columnwidth]{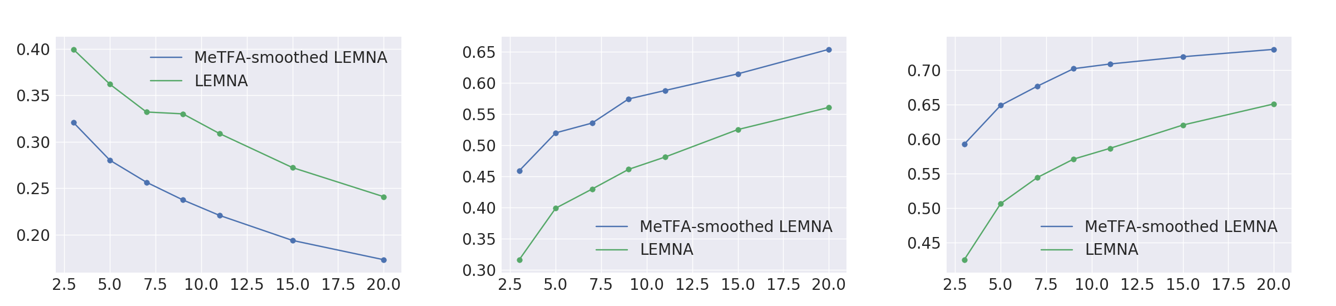}
     \caption{The results of of RFDT, RFAT and RST with different $n$ where $\mathbb{O}_n$ is $\mathbf{P}(0.7)$. From left to right, there are the results of RFDT, RFAT and RST, respectively. }
     \label{7_fidelity}
\end{figure}

\begin{figure}[htp]
     \centering
     \includegraphics[width=\columnwidth]{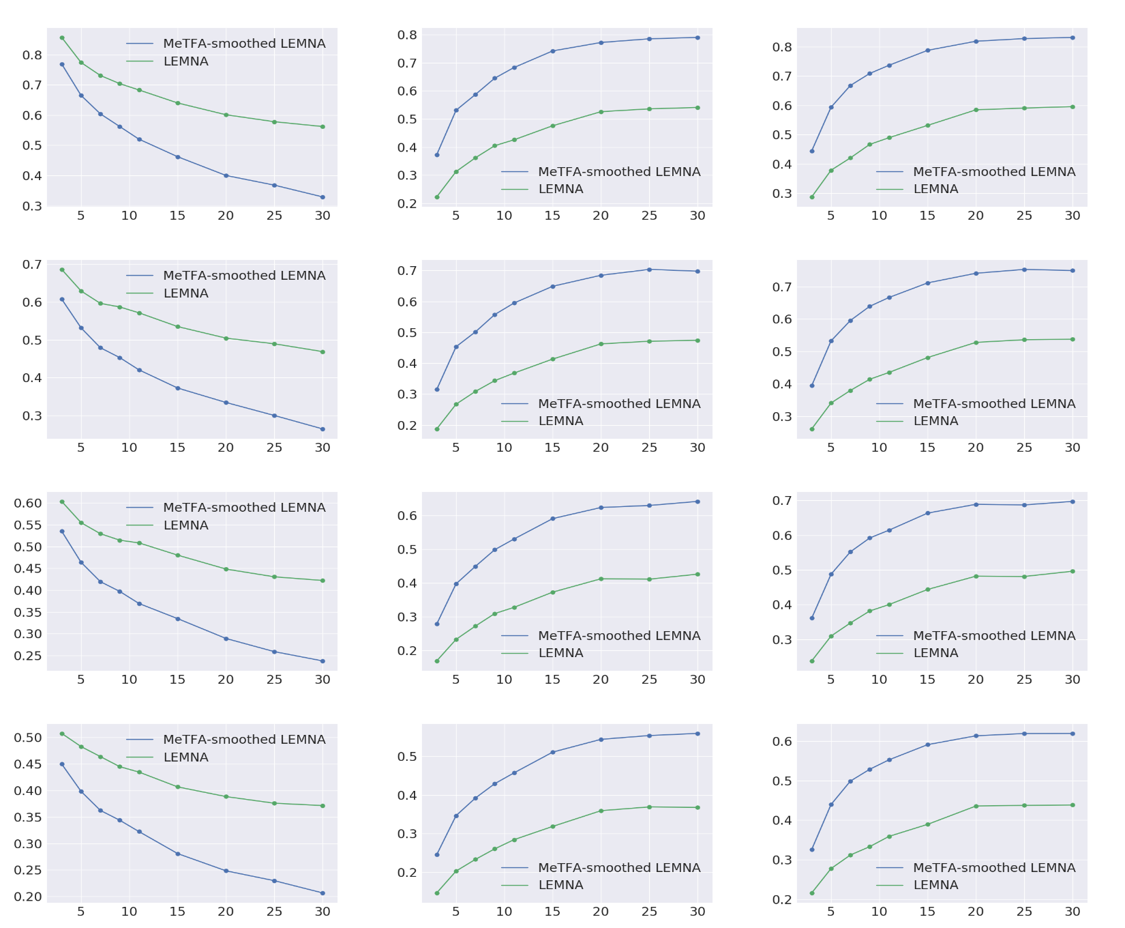}
     \caption{The results of Toxic Comment dataset where the number of samples for LEMNA is 500. The rows are the results of RFDT, RFAT and RST calculated with different $\mathbb{O}_n$. From top to bottom, $\mathbb{O}_n$ is $\mathbf{P}(0)$, $\mathbf{P}(0.3)$, $\mathbf{P}(0.5)$ and $\mathbf{P}(0.7)$, respectively. From left to right, the results are RFDT, RFAT and RST, respectively. }
     \label{fig: total_fidelity_toxic_500}
     % \vspace{-0.4cm}
\end{figure}

% \clearpage
\begin{figure}[htp]
     \centering
     \includegraphics[width=\columnwidth]{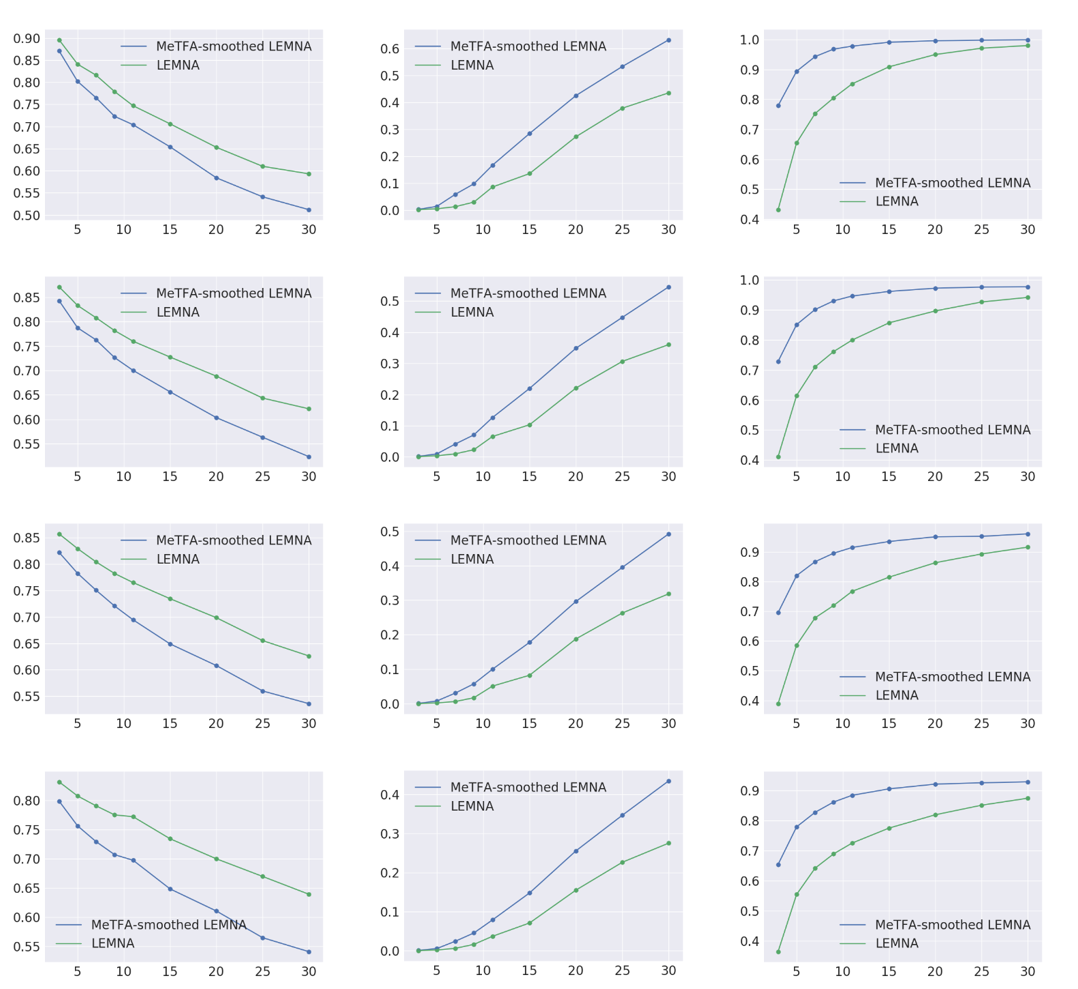}
     \caption{The results of IMDb Reviews dataset where the number of samples for LEMNA is 2000. The rows are the results of RFDT, RFAT and RST calculated with different $\mathbb{O}_n$. From top to bottom, $\mathbb{O}_n$ is $\mathbf{P}(0)$, $\mathbf{P}(0.3)$, $\mathbf{P}(0.5)$ and $\mathbf{P}(0.7)$, respectively. From left to right, the results are RFDT, RFAT and RST, respectively. }
     \label{fig:total_fidelity_sentiment_2000}
     % \vspace{-0.4cm}
\end{figure}

% \begin{figure}[htp]
% \centering
% \includegraphics[width=0.9\columnwidth]{figures/vs_png.png}
% \caption{Comparison of CAM and the MeTFA-smoothed CAM on 3 types of images. All of the explanations are generated with the predicted classes. The first rows in (a), (b) and (c) are generate with CAM, while the second rows in (a), (b)  and (c) are generated with the MeTFA-smoothed CAM. (a) is generated with clean images. (b) is generated with PGD-attacked images. (c) is generated with ADV$^2$-attacked images.}
% \label{vs_st}
% \end{figure}

% \clearpage
% \subsubsection{Algorithms}

\begin{algorithm}[ht]
  \Parameter{number of sampled explanation for MeTFA $n$, vanilla explanation $E$,  sampled distribution $\mathbb{P}$, target model $M$, target class $c$, target data $I$, confidence level $\alpha$}
  \KwOut{MeTFA-significant map $m1$, MeTFA-smoothed map $m2$, higher bound map $m3$ and lower bound map $m4$}

  \tcc{line 1 to 7 sample explanations around $I$}
  expl\_list=[]\\
  \For{$i\gets0$ \KwTo $n$}{
    $inner\_noise\sim \mathbb{P}$\\
    $I_s = I + inner\_noise$\\
    $expl = E(M,I_s,c)$\\
    expl\_list.append(expl)
  }
  \tcc{line 8 to 15 generate $m1$}
  perform 1D clustering with all values in expl\_list and get break value $h$\\
  Compute $ct_j(h)$ for every feature $j$\\
  $k_1 = \arg \max_k \{p\text{-value of} \ (V_j\leq h) \leq \alpha \}$\\
  $k_2 = \arg \max_k \{p\text{-value of}  \ (V_j\geq h) \leq \alpha \}$\\
  if $ct_j(h)\leq k_1$, $m1_j=-1$\\
  if $ct_j(h)\geq k_2$, $m1_j=1$\\
  else, $m1_j=0$\\
  \tcc{line 15 to 19 generate $m2$, $m3$ and $m4$}
  $k_1,k_2$ are calculated as above\\
  for every feature $j$, sort the value of expl\_list\\
  $m2_j$ is the average of the values between the $k_1$ th value and $k_2$ th value of expl\_list$_j$\\
  $m3_j$ is the $k_2$ th value of expl\_list$_j$\\
  $m2_j$ is the $k_1$ th value of expl\_list$_j$\\
  \textbf{Return} $m1$, $m2$, $m3$, $m4$
\caption{Generate MeTFA maps.}
\label{alg:MeTFA maps}
\end{algorithm}

\begin{algorithm}[ht]
  \Parameter{$n$ is the number of sampled explanation for MeTFA, $k$ is the number of insertion for RI, the vanilla explanation $E$, the outer noise distribution $\mathbb{O}_n$, the inner noise distribution $\mathbb{P}$, the test dataset $\mathbf{D}$, the target model $M$, the target class $c$}
  \KwOut{RI}
  \For{image I in $\mathbf{D}$}{
  \tcc{line 2 to 11 generate the MeTFA-smoothed explanation}
  expl\_list=[]\\
  \For{$i\gets0$ \KwTo $n$}{
    $inner\_noise\sim \mathbb{P}$\\
    $I_s = I + inner\_noise$\\
    $expl = E(M,I_s,c)$\\
    expl\_list.append(expl)
  }
  $k_1,k_2$ are calculated from Theorem 3\\
  for every pixel, sort the value of expl\_list\\
  for every pixel, take the average of the values between the $k_1$ th value and $k_2$ th value and get MS\_expl\\
  \tcc{line 12 to 17 calculate the RI metric}
  \For{$i\gets0$ \KwTo $k$}{
    $outer\_noise \sim\mathbf{O}_n$\\
    $I_n = I + outer\_noise$\\
    insertion = Insertion(MS\_expl, $I_n$, $M$, $c$)
  }
  \tcc{use the average of k insertion as the approximation of the expectation}
  take the average of $k$ insertion values to get RI value
  }
  \textbf{Return} the average of RI values
\caption{The RI calculation for MeTFA-smoothed explanation.}
\label{alg:calculation RI}
\end{algorithm}

\end{document}